\documentclass[10pt,twocolumn,letterpaper]{article}

\usepackage[pagenumbers]{cvpr} 

\usepackage{graphicx}
\usepackage{amsmath}
\usepackage{amssymb}
\usepackage{booktabs}

\usepackage{url,wrapfig}
\usepackage{caption}
\usepackage{graphicx}
\usepackage{xcolor}
\usepackage{amsmath}
\usepackage{setspace}
\usepackage{multirow}
\usepackage{wrapfig}
\usepackage{bm}
\usepackage{bbm}
\usepackage{enumitem} 
\usepackage{afterpage} 
\usepackage[accsupp]{axessibility}  

\usepackage{xspace}
\usepackage{comment}

\usepackage[pagebackref,breaklinks,colorlinks]{hyperref}
\usepackage[numbers,sort&compress]{natbib}
\usepackage{algorithm} 
\usepackage{algorithmic} 
\usepackage{mathtools} 

\usepackage{amsthm, amssymb, amscd, amsfonts} 
\usepackage{thmtools}
\usepackage{thm-restate} 
\usepackage{wrapfig}
\usepackage{stfloats}  

\newif\ifshowcomments 

\showcommentsfalse

\DeclareMathOperator*{\argmax}{arg\,max\,}

\newcommand{\one}{\mathbbm{1}}

\newtheorem{´desiderata}{Desiderata}

\providecommand{\datasetname}{StarCraftImage\xspace}
\providecommand{\dataseturl}{\href{https://starcraftdata.davidinouye.com/}{https://starcraftdata.davidinouye.com/}\xspace}

\renewcommand{\ie}{i.e.,\xspace}
\renewcommand{\eg}{e.g.,\xspace}

\renewcommand{\th}{\textsuperscript{th}\xspace}

\DeclarePairedDelimiter{\floor}{\lfloor}{\rfloor}

\providecommand{\argmax}{\mathop{\mathrm{argmax}}}

\usepackage[capitalize]{cleveref}
\crefname{section}{Sec.}{Secs.}
\Crefname{section}{Section}{Sections}
\Crefname{table}{Table}{Tables}
\crefname{table}{Tab.}{Tabs.}

\def\cvprPaperID{9987} 
\def\confName{CVPR}
\def\confYear{2023}

    \makeatletter
\def\@fnsymbol#1{\ensuremath{\ifcase#1\or \ddagger\or \dagger\or
   \mathsection\or \mathparagraph\or \|\or **\or \dagger\dagger
   \or \ddagger\ddagger \else\@ctrerr\fi}}
    \makeatother 

\begin{document}

\title{StarCraftImage: A Dataset For Prototyping Spatial Reasoning \\ Methods For Multi-Agent Environments}

\author{Sean Kulinski$^\dagger$ \\Purdue University
\and
Nicholas R. Waytowich \\ARL\thanks{DEVCOM Army Research Laboratory}
\and
James Z. Hare \\ARL$^\ddagger$
\and
David I. Inouye\thanks{Corresponding Authors: Sean Kulinski \href{mailto:smkulinski@gmail.com}{smkulinski@gmail.com} and David I. Inouye \href{mailto:dinouye@purdue.edu}{dinouye@purdue.edu}. \\ This work appeared in the Proceedings of the IEEE/CVF Conference on Computer Vision and Pattern Recognition (CVPR), 2023, pp. 22004-22013} \\Purdue University
}

\maketitle

\begin{abstract}
Spatial reasoning tasks in multi-agent environments such as event prediction, agent type identification, or missing data imputation are important for multiple applications (e.g., autonomous surveillance over sensor networks and subtasks for reinforcement learning (RL)). StarCraft II game replays encode intelligent (and adversarial) multi-agent behavior and could provide a testbed for these tasks; however, extracting simple and standardized representations for prototyping these tasks is laborious and hinders reproducibility. In contrast, MNIST and CIFAR10, despite their extreme simplicity, have enabled rapid prototyping and reproducibility of ML methods. Following the simplicity of these datasets, we construct a benchmark spatial reasoning dataset based on StarCraft II replays that exhibit complex multi-agent behaviors, while still being as easy to use as MNIST and CIFAR10. Specifically, we carefully summarize a window of 255 consecutive game states to create 3.6 million summary images from 60,000 replays, including all relevant metadata such as game outcome and player races. We develop three formats of decreasing complexity: Hyperspectral images that include one channel for every unit type (similar to multispectral geospatial images), RGB images that mimic CIFAR10, and grayscale images that mimic MNIST. We show how this dataset can be used for prototyping spatial reasoning methods. All datasets, code for extraction, and code for dataset loading can be found at \dataseturl.

\end{abstract}

\vspace{-1 em}
\section{Introduction}

\begin{figure}[!ht]
    \centering
    \includegraphics[width=1.005\columnwidth]{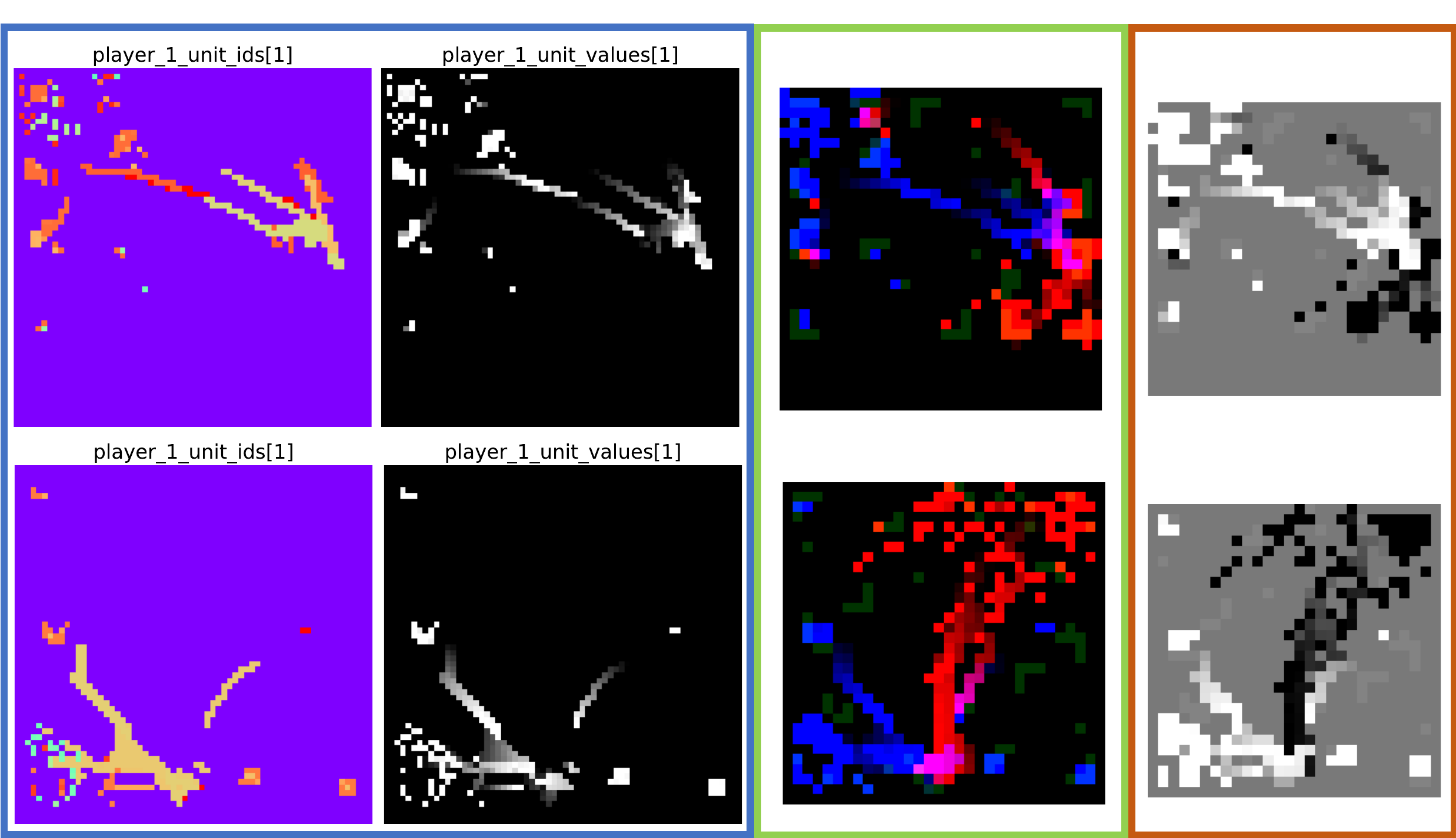}
    \vspace{-1.5em}
    \caption{Two samples (one per row) showing (Blue box/left) our 64 x 64 StarCraftHyper dataset which contains all unit IDs and corresponding values for both players (color for unit IDs denotes categorical unit ids), (Green box/middle) StarCraftCIFAR10 (32 x 32) which is easy to interpret where blue is player 1, red is player 2, and green are neutral units such as terrain or resources, and (Orange box/right) StarCraftMNIST (28 x 28) which are grayscale images further simplified to show player 1 as light-gray, player 2 as dark-gray, and neutral as medium-level shades of gray.}
    \vspace{-1.5 em}
    \label{fig:intro-dataset-examples}
\end{figure}

Spatial tasks in multi-agent environments require reasoning over both agents' positions and the environmental context such as buildings, obstacles, or terrain features.
These complex spatial reasoning tasks have applications in autonomous driving, autonomous surveillance over sensor networks, or reinforcement learning (RL) as subtasks of the RL agent.
For example, to predict a car collision, an autonomous driving system needs to reason about other cars, road conditions, road signs, and buildings.
For autonomous surveillance over sensor networks, the system would need to reason over the positions of objects, buildings, and other agents to determine if a new agent is normal or abnormal or to impute missing sensor values.
An RL system may want to predict the cumulative or final reward or impute missing values given only an incomplete snapshot of the world state, i.e., partial observability.
Yet, collecting large realistic datasets for these tasks is expensive and laborious.

Due to the challenge of collecting real-world data, practitioners have turned to (semi-)synthetic sources for creating large clean datasets of photo-realistic images or videos \cite{fabbri2021motsynth, geiger2012we, wang2019learning, krahenbuhl2018free}.
For example, \cite{fabbri2021motsynth} leveraged the Grand Theft Auto V game engine to collect a synthetic video dataset for pedestrian detection and tracking.
\cite{butenuth2011integrating} overlays aerial images with crowd simulations to provide a crowd density estimation dataset.
Yet, despite near photo-realism, these prior datasets focus on simple multi-agent environments (e.g., pedestrian-like simulations \cite{wang2019learning, fabbri2021motsynth}) and thus lack complex (or strategic) agent and object positioning.
In sharp contrast to these prior datasets, human-based replays of the real-time strategy game StarCraft II capture complex strategic and naturally adversarial positioning of agents and objects (e.g., buildings and outposts).
Indeed, the human player provides thousands of micro-commands that produce an overall intelligent and strategic positioning of agents and building units.
The release of the StarCraft II API and Python bindings \cite{vinyals2017starcraft} significantly reduces the barrier to using this rich data source for multi-agent environments.
Yet, the StarCraft II environment still requires significant overhead including game engine installation, looping through the game engine, understanding the API, etc.
This greatly limits the broad adoption of this very rich source of multi-agent interactions as a benchmark dataset---in contrast to the classic and extremely easy-to-use MNIST \cite{lecun1998mnist} and CIFAR10 \cite{krizhevsky2009learning} benchmark datasets 
that drove image classification research in the early years and continue to be used for prototyping new ML methods.
In summary, prior multi-agent datasets either lack complex strategic behavior or require significant implementation overhead.

To address these issues, we created \datasetname: a simplified image-based representation of human-played StarCraft II matches to serve as a large-scale multi-agent spatial reasoning benchmark dataset that is as easy to use as MNIST and CIFAR10 while still exhibiting complex and strategic object positioning.
As seen in \autoref{fig:intro-dataset-examples}, each image in \datasetname is akin to a detailed snapshot of the StarCraft II minimap and includes the locations of all units (both moveable units and buildings), the units' IDs, as well as important metadata like which player won that match, player resource counts, the current map name, player ranking, etc.
We made two key design decisions when developing \datasetname.
First, we chose to represent the matches by snapshot images that summarize a window of approximately 10 seconds of gameplay rather than a video.
This design choice was motivated both by the ease-of-use criteria (as images are easier to load and manipulate than videos) and by the goal of performing \emph{spatial} rather than temporal reasoning tasks---though a video dataset for complex temporal reasoning is a natural direction for future work.

Our second design choice was to represent the matches via minimap-like images rather than photo-realistic renderings of the game state.
This choice was motivated by two reasons.
First, minimap images are easy to use because they are small yet still represent of the whole environment.
By using a minimap representation, we can encode the most crucial game information (unit types, recent troop movements, building locations, environmental features, etc.) in a naturally compact representation.
Indeed, the minimap representation is critical for playing StarCraft II as evidenced by the following quote from the famous StarCraft II player Day[9] (Sean Plott): ``...the two most important things [are] the minimap and your money'' \cite{plott2010mental}.
Second, the minimap representation allows for us to have many diverse samples while still maintaining a small data footprint.
The resultant smaller disk size allows for rapid prototyping via quick dataset downloads and swift data consumption.
Compared to prior common spatial reasoning datasets, our proposed 3.6 million image dataset has a total disk size of 10.6Gb while the MOTSynth-MOT-CVPR22 dataset \cite{fabbri2021motsynth}, which consists of 1.3 million images, has a disk size of 167Gb  (16x larger, while containing half the number of samples).
Ultimately, we construct three different image representations with decreasing complexity: Hyperspectral images which give precise game state information by encoding the unit ids and last-seen timestamps at each spatial location (mimicking the hyperspectral geospatial representations), RGB images that mimic CIFAR10, and grayscale images that mimic MNIST.
Thus, our dataset is compatible with common ML frameworks with minimal overhead or preprocessing effort.
Overall, we use 60k StarCraft II replays to create 3.6 million summary images (not multi-counting different representations) and corresponding metadata.

To demonstrate how multi-agent spatial reasoning tasks can be easily prototyped using \datasetname, we also provide a series of benchmark tasks.
We perform target identification (i.e., determining unit type from only knowing unit locations) where the input is either an RGB or grayscale image and the target image is hyperspectral with each channel corresponding to a unit type.
We also perform more complex tasks such as map event prediction (i.e., game outcome and StarCraft race prediction) which serve as canonical image-level reasoning problems.
To show how our image representations can be easily manipulated for other tasks (like Rotated MNIST \cite{larochelle2007empirical} or Color MNIST \cite{arjovsky2019invariant}), we map missing data imputation as an image inpainting task using both simulated sensor network faults and the fog-of-war from the game engine.
Ultimately, we hope to provide a large-scale and rich multi-agent spatial reasoning dataset that is very easy to use yet exhibits complex and strategic placement of agents for complex spatial reasoning applications.
We summarize our contributions as follows:
\begin{itemize}
\setlength{\itemsep}{0em}
    \item We design and extract \datasetname as an easy-to-use multi-agent spatial reasoning dataset under three representations: 1) Hyperspectral images that encode all unit ids and lasts seen timestamps for each spatial location, 2) RGB images that mimic CIFAR10, and 3) grayscale images that mimic MNIST.
    \item We apply \datasetname on tasks such as target identification, movement prediction, and more. We also propose several noise simulation models and discuss several task modifiers such as domain generalization.
    \item We publicly release the datasets with a permissive CC BY 4.0 license.
    We also release the StarCraft II dataset extraction code and the relevant data loaders and modules for using the data as a Python package with an MIT license, and provide matainance as laid out in our dataset nutrition label: \autoref{tab:Dataset-Nutrition-Label}.
\end{itemize}

\vspace{-1 em}
\section{Dataset Extraction and Construction}
In this section, we describe how we extract observational data from the simulated yet complex environment of the StarCraft II (SC2) game. We then transform the raw data into the hyperspectral, CIFAR10, and MNIST formats that are readily usable in ML tools.

\subsection{Extracting Raw Data From SC2 Replays}
Due to SC2 being an almost entirely deterministic game, 
an SC2 replay file contains an entire list of actions from both players that can be used to re-simulate an entire match by passing the actions back to the SC2 game engine.
Each replay file also contains metadata from the match such as: the length of the match, the map/arena the match took place in, and per-player statistics such as the match making rating (MMR) (which can be thought of as the skill level of that player), the actions-per-minute (APM) the player took, and whether that player won, lost, or tied the match.
Additionally, Activison Blizzard (the maker of SC2) bundles large sets of these replays together as a Replay Pack for others to use. 
We used Replay Pack \texttt{3.16.1 - Pack 1} from \cite{s2client2015}.

\begin{figure*}[ht!]  
    \centering
    \includegraphics[width=0.9\textwidth]{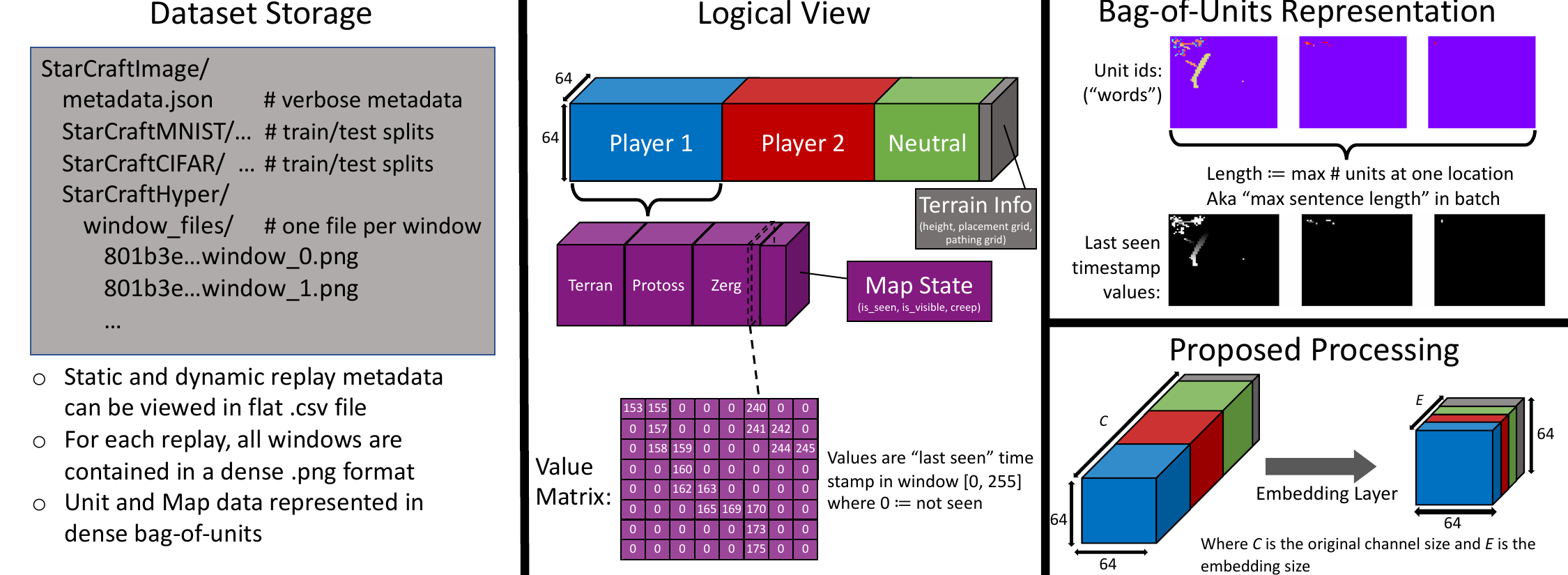}
    \vspace{-0.5em}
    \caption{An overview of our hyperspectral dataset from different perspectives.
    The raw image data is stored in texttt{.png} files using the bag-of-units representation.
    A logical view of the dataset is a (sparse) hyperspectral image with many channels that include unit information and visibility per player, resource information (neutral units), and map information. 
    The bag-of-units representation enables processing this very high-dimensional dataset using dense matrices only and leveraging embedding layers that are often used for processing sequences of IDs; importantly, because the unit order does not matter, an order-invariant reduction such as max or sum should be used to arrive at a representation with a fixed number of embedding channels $E$.}
    \vspace{-1 em}
    \label{fig:hyperspectral-overview}
\end{figure*}

To extract the game state, we used the \texttt{PySC2} \cite{vinyals2017starcraft} Python library developed for RL applications that interfaces with the SC2 game engine.
\texttt{PySC2} exposes the raw game state while re-simulating a match based on replay files.
Each raw game state consists of information such as the location, allegiance, size, unit type ID, and health of every unit (character, building, worker, solider, etc.) which currently exists for that specific frame (where a frame is a single unit of time in a game).
The raw frame data also contains dynamic map information including the visibility for each player (the locations on the map that the player can see due to friendly units/scouts being in that area, versus areas which are undiscovered and thus hidden) and the current creep state (which is a terrain feature consisting of purple slime in which most Zerg structures must be built and upon which Zerg units will move faster).
However, since the PySC2 interface was designed for \emph{interacting} with StarCraft II, it comes with a steep learning curve and a complex data representation -- which greatly hinders our goal of having clean observed game states that can be represented in a standard form.
Thus, we use PySC2 to extract raw game state observations and process these into standard image formats.

\subsection{StarCraftHyper: Construction and Processing of Hyperspectral Representation}
\label{ssec:sensorhyper}

Our most general format is a hyperspectral image format where each channel represents information for each unit type for each player in SC2.
To do this, we first use PySC2 to  extract raw frame data and for the $f$\th frame observation, we record the location of each unit present via $H_{f}[u_{PID}, x_h, y_h] = \one(u_{PID}, x_h, y_h)$ where $\one$ is an indicator function that returns 1 if a unit is present, else 0, $u_{PID}$ is the player-specific unit ID (PID), and $x_h,y_h$ is the spatial location of the unit.
Since the raw data gives spatial information in raw game-map spatial coordinates, we must perform a coordinate transform to our square hyperspectral image coordinates: $(x_{h}, y_{h}) = \floor*{\frac{(x_{raw}, y_{raw})}{\max(x_{raw}, y_{raw})}}$.
We also crop to only the playable area of the map.
There are 170 unit IDs for Player1, 170 IDs for Player2 units, and 44 IDs for Neutral units, which is 384 $PID$s (i.e., channels).

Next, to allow for video-like spatial movements, we form a stack of 255 consecutive hyperspectral images $H_{stack} = [H_f]$ where $f \in [0, 255]$.\footnote{We chose 255 to ensure that the values fit in an unsigned 8-bit integer and captures roughly 10 seconds of real time.}
Given $H_{stack}$, which is a tensor with shape $(255, 384, 64, 64)$, we simplify this from a video format to a static image format by collapsing the time axis to create the summarized hyperspectral image $H$.
We do this by recording the frame index of the most recent frame where a unit was present for each $(PID, x, y)$ coordinate (\ie $H[c, x, y] = \argmax_f H_{stack}[f, c, x, y] \neq 0$) and if $H_{stack}[f, c, x, y] = 0 ~ \forall f \in [0, 255]$, then $H[c, x, y] = 0$).
Another possibility would be to average over the window of frames instead of the last seen timestamp; however, the last seen timestamp enables simple visualization of movement via a ghosting-like effect, and the last seen timestamp preserves time information (albeit only a compressed amount).

\begin{figure*}[!ht]
    \centering
    \includegraphics[width=0.9\textwidth]{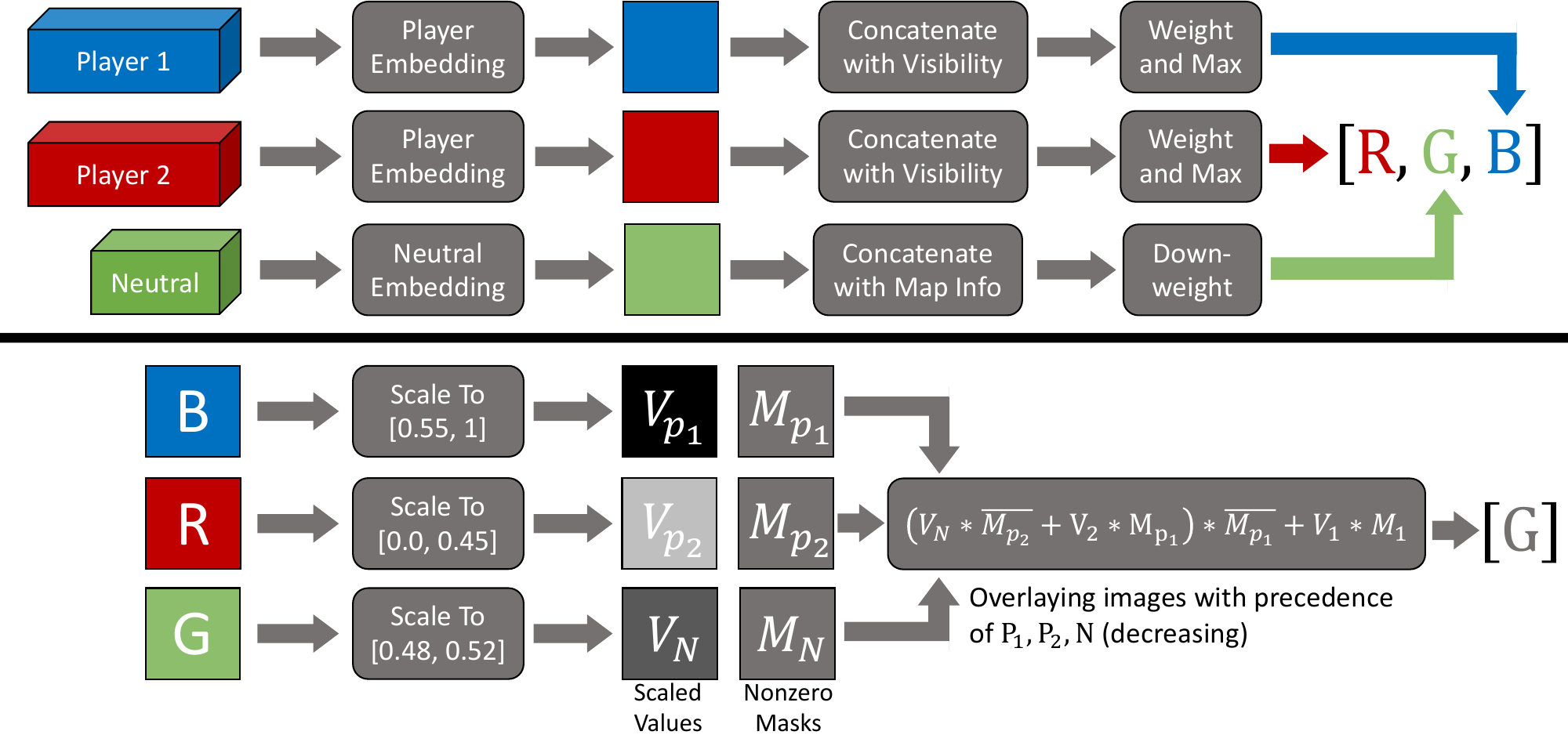}
    \vspace{-1 em}
    \caption{(Top) We embed the unit information of player 1, player 2, and neutral separately using an embedding of size 1. We then combine with other dense features (visibility for players and terrain info for neutral).
    Finally, we concatenate each output into a 3-channel 32x32 px RGB image where the neutral channel is down-weighted for visual clarity.
    (Bottom) We take the RGB color image, rescale the values of each channel, and overlay each channel into a single grayscale 28x28 px image where precedence is given to P1, then P2, and finally neutral or background. 
    We use precedence combinations as linear combinations of the layers could lead to unit information being canceled.
    }
    \vspace{-1 em}
    \label{fig:rbg-grayscale-overview}
\end{figure*}

While this condensed representation does have the trade-off that if a unit with the same $PID$ crosses the same $(x,y)$ location more than once in a frame stack, only the last crossing will be recorded in $H$, this is rare and seems a reasonable trade-off for a much simpler representation. 
At this step, the non-zero entries of $H$ are saved as the raw representation of our dataset---i.e., a sparse tensor representation.
We can compress $H$ into a dense ``bag-of-units'' representation, which is similar to representing a sequence of words by their IDs rather than by very high-dimensional one-hot vectors, but where the order of the IDs does not matter (hence, the term ``bag'' as in bag-of-words representations).
Just as the number of words of a sentence can vary in NLP, the number of units at each location can vary.
Therefore, as in processing word sequences, we pad the channels of the dense representation with zeros (representing no unit) up to the max number of units at any location (either in a single sample or in a batch of samples), denoted by $k$.
Concretely, the bag-of-units representation collapses the channel axis into $k$ ID matrices and $k$ timestamp matrices of size $(64, 64)$, where the ID matrix contains the $PID$ of the units present at each $(x,y)$ coordinate, the timestamp matrices contain the corresponding timestamp that the unit was last seen, and $k$ is the max number of units present at one $(x,y)$ location in $H$.
This highly-compressed bag-of-units representation for the StarCraftHyper dataset can be seen in the top right of \autoref{fig:hyperspectral-overview} and is the default representation for the StarCraftHyper dataset.


\subsection{StarCraftCIFAR10 and StarCraftMNIST: RGB and Grayscale Representations}

To further simplify dataset usage and prototyping ability, we develop datasets that mimic CIFAR10 and MNIST in terms of image size, number of channels, number of classes, and number of train/test samples as seen in (middle) and (right) of \autoref{fig:dataset-examples}.
Thus, our StarCraftCIFAR10 and StarCraftMNIST datasets can be used for rapid initial prototyping of new spatial reasoning methods just as these ubiquitous datasets have been used for prototyping image classification.
These can model situations where agent and building positions are known but the agent type is unknown (e.g., low resolution satellite images or a network of pressure sensors).
One natural task is to infer unit types given only unit location information, which is discussed in more detail in future sections.

To construct StarCraftCIFAR10, we first follow the approach in \autoref{ssec:metadata-analysis} to subsample our StarCraftHyper dataset to 50,000 train windows and 10,000 test windows, to match the dataset size of CIFAR10.
To transform each hyperspectral window into a CIFAR10 format, we separate $H$ into player-specific images and follow the process shown in \autoref{fig:rbg-grayscale-overview} (top).
To construct the StarCraftMNIST dataset, we similarly subsample from the full StarCraftHyper dataset, but to a size of 60,000 train and 10,000 test images as in MNIST.
We process the images in the manner seen at the top and bottom of \autoref{fig:rbg-grayscale-overview}, where the last step is a function that overlays the $V_{p_1}, V_{p_2}, V_{N}$ scaled maps on top of each other such that any non-zero elements of $V_{p_2}$ will overwrite the nonzero elements of $V_{N}$ and nonzero $V_{p_1}$ values will overwrite both.
We decided to overwrite rather than average because having a unit of player 1 and player 2 at the same location would average to a gray background value but that is in fact one of the most interesting locations.
In the next section, we discuss the creation of the 10 classes for each window via a combination of the variables: Player 1 race, Player 2 race, and Player 1 outcome.


\subsection{Dataset Exploration and Analysis}
\label{ssec:metadata-analysis}

\begin{figure*}[!ht]
    \centering
    \includegraphics[width=.95\textwidth]{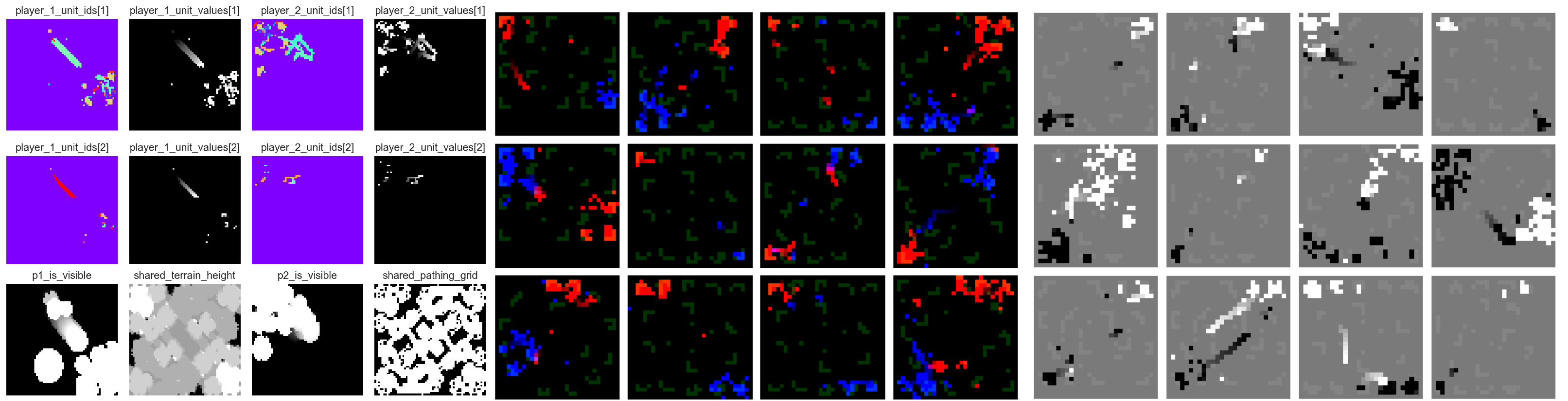}
    \vspace{-0.5 em}
    \caption{(Left) Our 64 x 64 StarCraftHyper dataset contains all unit IDs and corresponding values for both players (color for unit IDs denotes categorical unit ids) where visibility is player specific but the terrain and pathing grid are shared (a few other layers are not shown, see appendix). (Middle) StarCraftCIFAR10 (32 x 32) is easy to interpret where blue is player 1, red is player 2, and green are neutral units which are usually just resources. (Right) MNIST (28 x 28) grayscale images are further simplified to show player 1 as white to white-gray, player 2 as black to black-gray, and neutral as shades of gray.}
    \label{fig:dataset-examples}
    \vspace{-1 em}
\end{figure*}

All in all, the \datasetname dataset consists of 3,607,787 windows extracted from 60,000 replays which are readily available in three representations (examples in \autoref{fig:dataset-examples}). 
The image data for each window is stored as a \texttt{.png} file in the bag-of-units representation.
The data can be accessed via directly loading in the relevant \texttt{.png} file and metadata row, or more simply by using the corresponding PyTorch dataset classes that we have developed (one class for each representation).

Jointly with the image data collection, we also aggregated relevant metadata for each window, such as the temporal location of the window in the overall match (\eg 75\th window of 130), which player won the match, the races of the players, the name of match's map, etc. (for a full list of the metadata keys, please see \autoref{sec:metadata-description}).
This metadata has many uses for filtering replays based on conditions for a specific application, \eg training on a subset of maps and testing on the held out set.
Additionally, for canonical class labels, we use the race of each player (Terran, Zerg, or Protoss) and player 1's outcome (Win and NotWin where NotWin includes the rare Tie outcome) to split the overall dataset into 18 classes (3 races for player 1, 3 races for player 2 and 2 outcomes).
We chose these three variables (Player 1 outcome, Player 1 race, Player 2 race) because though outcome prediction is a canonical task, readily available ground truth for race prediction with this dataset is akin to behavior or tactical strategy prediction, as unit type information is hidden in the StarCraftCIFAR10 and StarCraftMNIST versions of the dataset.
For StarCraftCIFAR10 and StarCraftMNIST, we select only classes that have at least one player as Zerg (5 total) with both outcomes to get exactly 10 balanced classes to match the setup of CIFAR10 and MNIST---this could be done similarly for Terran and Protoss but Zerg is the easiest to understand because some Zerg-specific units are often spread across the battlefield.

\section{Multi-Agent Spatial Reasoning Applications}
\label{sec:ml-tasks-on-sn}
In this section, we list examples of spatial reasoning tasks on our datasets (e.g., global reasoning as a classification task). We will also discuss simple noise models that simulate more complex scenarios on top of the clean data representations. 
Finally, we discuss natural task modifiers such as domain generalization or adversarial contexts.
In all cases, we aim for a compromise between realism and simplicity as this dataset is meant as an initial prototyping dataset for complex or strategic agent and object positioning rather than a fully realistic spatial reasoning dataset.
Given space constraints, we provide demos of these tasks in the supplementary material both in \autoref{sec:additional-experiments-APPENDIX} and as IPython notebooks in our code repository.

\subsection{Spatial Reasoning Examples}

\paragraph{Target identification (Image colorization)}
The goal here is to identify the unit type (e.g., marine unit) or affiliation (player one, two or neutral) for every detected unit.
This can be seen as a setting where an image only shows if a unit exists in its field of view (e.g., an aerial photo from a UAV or a post-processed output from a LiDAR scanner).
For the task, we cast this problem as an image colorization problem in which the input is either a StarCraftCIFAR10 or StarCraftMNIST and the target output is the corresponding StarCraftHyper or StarCraftCIFAR10 image.
   
\paragraph{Movement prediction (Simplified Multi-Object Tracking)}
Predicting what is going to happen next is clearly an important task especially in time-critical applications such as autonomous driving \cite{geiger2012we}, disaster relief \cite{AHRSHAS2013}, or, more generally, optical flow \cite{baker2011database}. 
While we do not generally consider the time dimension after we summarize the window, for this task, we can use the metadata to create pairs of adjacent window summary images where the input is the current summary image and the target output is the next summary image in the same match.

\paragraph{Predict final outcome or race (Classification)}
Spatial reasoning systems are often used to predict the global properties of a system (e.g., crop yield predictions \cite{masjedi2020prediction} or reward predictions for RL models \cite{vinyals2017starcraft}), which can be cast as classification.
The most canonical task is to predict the final outcome of the game (i.e., which player will win), which requires reasoning over both fighting units and environmental factors such as buildings and resources (\eg even if there is little movement/few fighting units in a window, a model can still predict who will win based off of who has the strongest base).
Another canonical task for the simplified datasets StarCraftCIFAR10 and StarCraftMNIST (which give only unit location information rather than unit type information) is to predict both players' races, which requires recognizing the common placement configurations for each race.

\paragraph{Imputing missing data (Image inpainting)}
Another critical task in spatial reasoning is imputing missing values for areas that lack coverage due to occlusion, data collection failures, or adversarial attacks.
~\cite{guo2011sparsity, wang2020robust}.
Here the input image is a corrupted version of a sample from one of the three datasets, and the target output image is the uncorrupted sample.
Due to \datasetname's simple minimap representation, simulating spatial corruptions (\eg noisy measurements or partial observability) is simple to do--unlike in photo-realistic settings which would require editing the images or videos to hide or remove information.
In the next section, we go over examples of spatial corruption models.

\begin{figure*}[ht!]
    \centering
    \includegraphics[width=\textwidth]{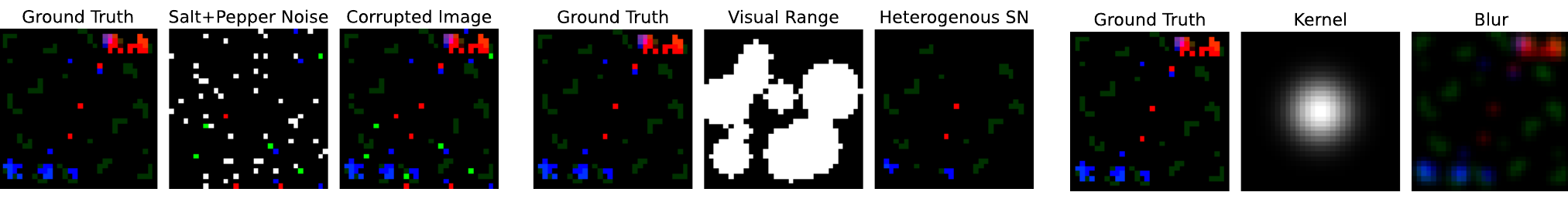}
    \vspace{-2 em}
    \caption{Three example noise corruption models which are simulated on top of the StarCraftCIFAR10 dataset, where (left) simulated random additive noise, (middle) simulates observations via a heterogeneous SN, and (right) simulates limited precision (blurry) observations.
    }
    \label{fig:simulated-sn-models}
    \vspace{-1.5 em}
\end{figure*}

\subsection{Simulated Data Corruption Models}
\label{ssec:simulated-sn-models}

\paragraph{Random additive noise}
This corruption model is relevant for settings where images are taken using noisy equipment or a hierarchical system where reasoning happens on a (potentially noisy) abstracted spatial representation. 
We can implement this as a type of salt and pepper noise where the salt noise can randomly add units to locations that do not have units (i.e., false positives) and each real unit could possibly become missing (i.e., false negatives), as seen in the left of \autoref{fig:simulated-sn-models}.


\vspace{-0.5 em}
\paragraph{Heterogeneous partial observations (Image masking)}
Here the images can be seen as the fusion of irregular heterogeneous sensor networks.
This can be simulated by producing a mask that is based on static sensor locations and detection ranges, see \autoref{fig:simulated-sn-models} (middle) for example.
Furthermore, detailed sensor models can be used to pre-process the masked observations to provide an accurate representation based on the type of sensor implemented at a particular location, e.g., acoustic sensors may only return a range of the unit relative to its position.
Sensor faults as above could be implemented on top of this heterogeneous sensor network (\eg masking over a set of sensors' visible range).
For examples and benchmark results on such heterogeneous sensor placements with aggregation failure simulations, please see \autoref{sec:benchmarks-on-sn-tasks-APPENDIX}.

\vspace{-0.5 em}
\paragraph{Imprecise sensors (Blur)}
Low resolution imaging will yield imprecise unit locations.
Thus, we can implement this noising process by performing blur operations on top of the original datasets.
This corruption is simplest to apply to StarCraftCIFAR10 (\eg \autoref{fig:simulated-sn-models}, right) and StarCraftMNIST via standard CV packages but could also be applied to StarCraftHyper (albeit with more computation).

\subsection{Spatial Reasoning Task Modifiers}
\paragraph{Robustness to distribution shift (Domain generalization)}
A key challenge in applying ML to real-world settings is training a model in one context but applying it to another context \cite{zhou2022domain}.
This is known as the domain generalization problem in which the goal is to perform the task well on an \emph{unseen} test domain \cite{quinonero2009dataset}.
The metadata that we provide can provide natural segmentations of the dataset into domains.
One of the most canonical examples of distribution shift in real-world settings is a change in
the environment settings ~\cite{koh2021wilds}. 
While greatly simplified, we can simulate changes in location by splitting the dataset based on the SC2 map and holding out one or more maps for testing.
Other excellent domain splits could be players' MMR or APM, which correspond to their skill level and frequency of actions.
Player two's race (Terran, Protoss, or Zerg) is also another way to split the dataset into 9 domains such as Terran vs. Terran, Protoss vs. Zerg, or Terran vs. Protoss.

\vspace{-0.5 em}
\paragraph{Robustness to adversarial attacks (Adversarial training)}
While uncommon, adversarial attacks are a genuine concern for reasoning methods, especially those which involve humans such as autonomous driving. 
We can simulate this idea by applying adversarial training methods under different adversarial attack models such as L0 pixel-wise attacks \cite{su2019one} for attacking individual units.
The adversarial training literature already benchmarks using MNIST as a key difficult example \cite{madry2017towards}, and thus, these StarCraft datasets could be immediately relevant and provide a more realistic benchmark for the adversarial training literature.

\paragraph{Equipment usage optimization (Active learning)}
Optimizing sensor location and power usage are key challenges in sensor networks~\cite{HSGW2020,ESS2019}. 
Following the simulated sensor network seen in the previous section, constrained power usage could be framed as an active learning problem in which the algorithm can only query a fixed number of sensors for each prediction problem.
For optimizing sensor location, the algorithm could attempt to determine where to place the next sensor (i.e., to uncover information at a certain location) to optimize the downstream task such as outcome classification.
A more complex case is moving sensors from their original locations to another location under a budget on geographic movement (e.g., a sensor on a robotic device).

\section{Benchmark Evaluations}
While we point the reader to \autoref{sec:benchmarks-on-sn-tasks-APPENDIX}, where we give full descriptions and results, here we introduce four benchmark multi-agent spatial reasoning tasks, which incorporate training U-Net-based \cite{ronneberger2015u} ResNet \cite{he2016deep} models.
The four benchmark tasks consist of two tasks on target identification (given a 64x64 RGB image, predict the ID of each unit at each location) and two tasks for unit tracking (given hyperspectral window $k$, predict what will happen in window $k+1$).
Both task sets consist of first training and evaluating on ``clean'' (unaltered) data.
To highlight the extendability of \datasetname, we also perform both tasks on corrupted data that has been passed through a simulation of a noisy sensor network.
The sensor network simulation consists of 50 imaging sensors with a radius of 5.5 pixels with different sensor placement methodologies (e.g., grid, random) and communication failures during sensor fusion (see \autoref{fig:sensor-simulation-masks} for details), and results in noisy training windows.
From the results seen in \autoref{tab:benchmark-results-MAIN}, it is clear that this is a difficult problem, especially when reasoning over corrupted samples, and hopefully future work can build upon these results.

\begin{table}[]
\caption{Benchmark Evaluations on Unit Type Identification and Next Hyperspectral Window Prediction with clean data and simulated data corruptions.
}
\vspace{-0.5 em}
\label{tab:benchmark-results-MAIN}
\resizebox{\columnwidth}{!}{%
\begin{tabular}{lcccccc}
\hline
\multicolumn{1}{c}{} &
  \multicolumn{3}{c}{Unit Identification (Acc)} &
  \multicolumn{3}{c}{Next Wind. (MSE)} \\ 
\multicolumn{1}{r}{Placement $\Rightarrow$} &
  \multicolumn{1}{l}{Clean} &
  \multicolumn{1}{l}{Grid} &
  \multicolumn{1}{l}{Rand.} &
  \multicolumn{1}{l}{Clean} &
  \multicolumn{1}{l}{Grid} &
  \multicolumn{1}{l}{Rand.} \\ \hline
Unet-ResNet18       & 56.6\% & 40.3\% & 30.1\% & 3.97 & 4.11 & 4.15 \\ 
Unet-ResNet34       & 58.5\% & 40.2\%   & 30.8\%   & 3.99 & 4.12 & 4.17 \\ 
Unet-ResNet50       & 62.5\% & 44.0\%   & 32.8\% & 4.00 & 4.06 & 4.15 \\ 
\hline
\end{tabular}%
}
\end{table}


\section{Preliminary Real-World Experiment on DOTA Satellite-Image Dataset}
\begin{figure}[!h]
    \centering
    \includegraphics[width=\columnwidth]{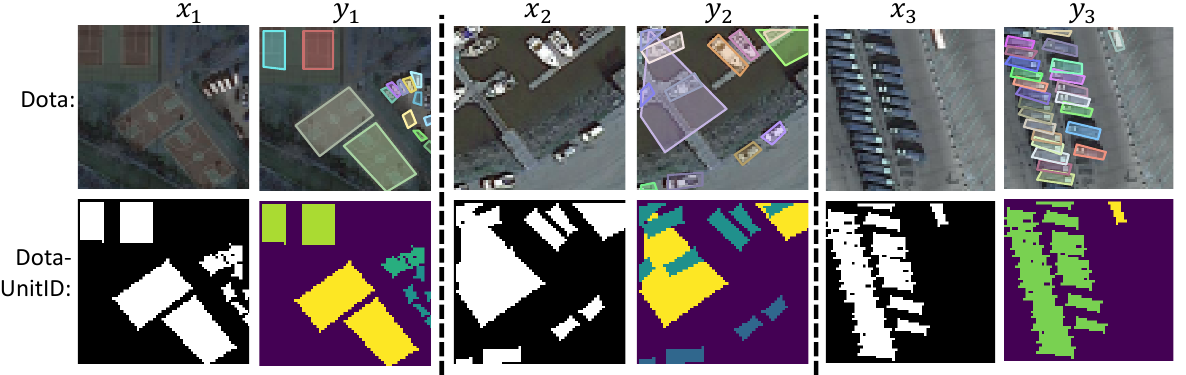}
    \caption{DOTA dataset examples, where the top row shows three original (input, annotations) pairs from the DOTA dataset \cite{Xia_2018_CVPR}, while the bottom row shows the three corresponding (input, label) pairs from our DOTA-UnitID dataset.
    The DOTA-UnitID task is to colorize the grayscale annotation mask.}
    \label{fig:dota-examples}
\end{figure}

In this section, we explore whether performance on StarCraftImage is predictive of performance on real-world datasets.
To this end, we use a version of the DOTA dataset \cite{Xia_2018_CVPR}, which is a benchmark dataset for multi-object detection in satellite images, where the samples have been transformed to match a similar format to StarCraftImage, which we call DOTA-UnitID (see \autoref{fig:dota-examples}).
This format is similar to the scenario when we may have remote sensing or a sensor network that can detect the presence of certain agents or buildings but may not know what they are (\eg due to cloud cover only synthetic aperture radar data is available).

In addition to the Unet-ResNet models seen above, we trained two state-of-the-art segmentation models, a SegFormer transformer model \cite{xie2021segformer} (12\th place in the CityScapes Test leaderboard \cite{cordts2016cityscapes}) and a Lawin transformer model \cite{yan2022lawin} (3\textsuperscript{rd} place in \cite{cordts2016cityscapes}) on the clean Unit Identification task for both the \datasetname dataset and the DOTA-UnitID datasets.
As seen in the second row of \autoref{tab:rebuttal-unit-id-results}, the model \emph{ranking} is the same for both the DOTA-UnitID and StarCraftImage-UnitID experiments across all models (e.g., the Unet-ResNet50 had the best unit accuracy across both datasets), thus providing preliminary evidence that performance improvements on our dataset will carry over to real-world datasets.
We note that the transformer results are much below the results of the ResNet models.
This is likely due to these larger models requiring longer training times than the CNN-based models.
Despite this, these results suggest that StarCraftImage is still a difficult dataset even for SOTA models.

\begin{table}[!h]
\caption{Unit-ID experiment results on clean data for StarCraftImage and Dota-UnitID. 
RX is short for a Unet-ResNet-X model.}
\vspace{-0.5 em}
\label{tab:rebuttal-unit-id-results}
\resizebox{\columnwidth}{!}{%
\begin{tabular}{l|lllll}
Model  & Lawin \cite{yan2022lawin} & SegFormer \cite{xie2021segformer}  & R18 & R34  & R50  \\ \hline
SCII   & 27.0\%  & 27.9\%  & 56.6\% & 58.5\% & 62.5\% \\
DOTA   & 34.1\%  & 35.0\%  & 52.4\% & 52.8\% & 53.6\%
\end{tabular}%
}
\end{table}

\vspace{-1.5 em}

\section{Related Works}

As with any ML task, accessible datasets are critical for making advancements.
For spatial reasoning tasks, these include elementary reasoning datasets (e.g., CLEVR \cite{johnson2017clevr}), scene understanding (e.g., Places \cite{zhou2017places}), geospatial datasets (e.g., Chesapeake Land Cover \cite{robinson2019large}), optical flow datasets (e.g., Middlebury \cite{baker2011database}), and more.

\vspace{-1 em}
\paragraph{Multi-Agent Spatial Datasets}
A notable area for multi-agent spatial reasoning tasks is reasoning for autonomous driving. 
For this, the well-known KITTI dataset \cite{geiger2012we} has driven many advancements since its introduction in 2012, and more recently the Waymo Open dataset\cite{sun2020scalability}) has introduced 1.1K additional scences with LiDAR and Camera measurements for practitioners to benchmark on.
More generally, there is TAO \cite{dave2020tao}, a multi-object tracking dataset, which is akin to a video-version of Microsoft COCO \cite{lin2014microsoft} and has over 800 object classes.
In a similar vein to our work exists pedestrian and crowd analysis (\eg crowd counting \cite{wang2019learning, chan2008privacy}, person ReID \cite{zheng2016person}, population density estimation \cite{butenuth2011integrating}), however, these datasets tend to have simple agent behvaiors such as conversing or walking from one point to another across a scene.

\begin{table}[ht]
\caption{An overview of multi-object spatial reasoning datasets.
\datasetname has the most complex agent positioning, the lowest overhead, and the ability to simulate more complex scenarios (\eg data corruption, as seen in \autoref{ssec:simulated-sn-models}).
GT stands for ``ground truth''.
}
\vspace{-0.5 em}
\label{tab:related-works}
\resizebox{\columnwidth}{!}{%
\begin{tabular}{lcllccc}
\hline
Dataset &
  \begin{tabular}[b]{@{}c@{}}Frame \\ Count\end{tabular} &
  Agent Positioning &
  \multicolumn{1}{c}{\begin{tabular}[b]{@{}c@{}}Over- \\ head\end{tabular}}  &
  \multicolumn{1}{c}{\begin{tabular}[b]{@{}c@{}}GT\end{tabular}} &
  \multicolumn{1}{c}{\begin{tabular}[b]{@{}c@{}}Noise \\ Sim\end{tabular}} \\ \hline
USD \cite{chan2008privacy}     & 2K   & Real Pedestrian    & Some &                       & \multicolumn{1}{c}{} \\
GCC \cite{wang2019learning}     & 15K  & Simulated Crowd &  Low &  \checkmark                    &                      \\
GTA \cite{krahenbuhl2018free}    & 250K & Simulated Ped/Drive   & Some &  \checkmark                    &                      \\
MOTSynth \cite{fabbri2021motsynth}\!\!\!\!\!\! & 1.4M & Simulated Pedestrian  & Some  & \checkmark                    &                      \\
TAO \cite{dave2020tao}     & 2.2M & Real YouTube    & Some &  \multicolumn{1}{l}{} &                      \\ \hline
\begin{tabular}[t]{@{}l@{}}PySC2\citep{vinyals2017starcraft} \\ \quad + Replays \end{tabular} &
  n/a &
  \multicolumn{1}{l}{\begin{tabular}[t]{@{}l@{}}Complex / Strategic \\(from human player)\end{tabular}}  &
  High &
  \checkmark &
  \\
\begin{tabular}[t]{@{}l@{}}\datasetname\!\!\!\!\!\!\\  (ours)\end{tabular} &
  3.6M &
  \multicolumn{1}{l}{\begin{tabular}[t]{@{}l@{}}Complex / Strategic \\(from human player)\end{tabular}}  &
  Low &
  \checkmark &
  \multicolumn{1}{c}{\checkmark} \\
  \hline
\end{tabular}%
}
\vspace{-1.5 em}
\end{table}

\vspace{-1 em}
\paragraph{Synthetic Datasets}
Developing multi-agent spatial reasoning datasets can be expensive as they tend to involve humans in the collection process. 
Thus, practitioners have turned to collecting this data from simulations of the real world.
For pedestrian tracking, there is the MOTSynth dataset \cite{fabbri2021motsynth}, GCC \cite{wang2019learning}, and the GTA dataset \cite{krahenbuhl2018free} which all use Grand Theft Auto V to produce realistic pedestrian images/behaviors as agents walk across a scripted scene.
Following \cite{krahenbuhl2018free}, the GTAV's rendering engine is used to produce exact crowd counts for \cite{wang2019learning} and bounding boxes, segmentation masks, and depth masks of all agents for \cite{krahenbuhl2018free, fabbri2021motsynth}.

\section{Conclusion}



We introduce \datasetname as a multi-agent spatial reasoning dataset with the overarching goal of being as easy to use for prototyping and initial method testing as MNIST and CIFAR10 while capturing complex and strategic unit positioning for advanced spatial reasoning methods.
To this extent, we process raw frame data from 60 thousand human StarCraft II replays to formulate 3.6 million summary images in three representations of decreasing complexity: StarCraftHyper which is hyperspectral images that encode the unit ids and last seen timestamps at each spatial location, StarCraftCIFAR10 which is RGB images that mimic CIFAR10, and StarCraftMNIST which is grayscale images that mimic MNIST.
We also include relevant metadata for each summary image which can be used to filter the \datasetname dataset when performing the tasks, corruption extensions, and modifiers we discuss in \autoref{sec:ml-tasks-on-sn}.
While we hope this work allows for easy prototyping and thus simpler and more systematic advances in developing spatial reasoning methods, we recognize that although this dataset is based on complex human actions, it is still a simplified simulated environment, and thus real-world data (or more \emph{realistic} data)  will always be needed to fully evaluate methods.
Additionally, our code for dataset processing, extracting, and loading the data could be used to expand or specialize new StarCraft datasets for multi-agent spatial reasoning applications using the millions of publicly available StarCraft II replays via Blizzard's developer API without the overhead of starting from scratch.
Ultimately, we hope our dataset provides the ML community with an easy-to-use multi-agent spatial reasoning dataset that will significantly reduce the barrier of entry for these important tasks.

\vspace{-0.5 em}
\paragraph{Acknowledgements} This work was supported by NSF (IIS-2212097) and ARL (W911NF-2020-221).

\clearpage

{\small
\bibliographystyle{ieee_fullname}
\bibliography{seank-ref}

\begin{thebibliography}{10}\itemsep=-1pt

\bibitem{AHRSHAS2013}
Naveed Ahmad, Mureed Hussain, Naveed Riaz, Fazli Subhani, Sajjad Haider, Khurram~S Alamgir, and Fahad Shinwari.
\newblock Flood prediction and disaster risk analysis using gis based wireless sensor networks, a review.
\newblock {\em Journal of Basic and Applied Scientific Research}, 3(8):632--643, 2013.

\bibitem{arjovsky2019invariant}
Martin Arjovsky, L{\'e}on Bottou, Ishaan Gulrajani, and David Lopez-Paz.
\newblock Invariant risk minimization.
\newblock {\em arXiv preprint arXiv:1907.02893}, 2019.

\bibitem{baker2011database}
Simon Baker, Daniel Scharstein, JP Lewis, Stefan Roth, Michael~J Black, and Richard Szeliski.
\newblock A database and evaluation methodology for optical flow.
\newblock {\em International journal of computer vision}, 92(1):1--31, 2011.

\bibitem{butenuth2011integrating}
Matthias Butenuth, Florian Burkert, Florian Schmidt, Stefan Hinz, Dirk Hartmann, Angelika Kneidl, Andr{\'e} Borrmann, and Beril Sirmacek.
\newblock Integrating pedestrian simulation, tracking and event detection for crowd analysis.
\newblock In {\em 2011 IEEE International Conference on Computer Vision Workshops (ICCV Workshops)}, pages 150--157. IEEE, 2011.

\bibitem{chan2008privacy}
Antoni~B Chan, Zhang-Sheng~John Liang, and Nuno Vasconcelos.
\newblock Privacy preserving crowd monitoring: Counting people without people models or tracking.
\newblock In {\em 2008 IEEE conference on computer vision and pattern recognition}, pages 1--7. IEEE, 2008.

\bibitem{s2client2015}
P.W.D. Charles.
\newblock S2clinet-proto repository.
\newblock \url{https://github.com/Blizzard/s2client-proto/\#replay-packs}, 2015.

\bibitem{cordts2016cityscapes}
Marius Cordts, Mohamed Omran, Sebastian Ramos, Timo Rehfeld, Markus Enzweiler, Rodrigo Benenson, Uwe Franke, Stefan Roth, and Bernt Schiele.
\newblock The cityscapes dataset for semantic urban scene understanding.
\newblock In {\em Proceedings of the IEEE conference on computer vision and pattern recognition}, pages 3213--3223, 2016.

\bibitem{dave2020tao}
Achal Dave, Tarasha Khurana, Pavel Tokmakov, Cordelia Schmid, and Deva Ramanan.
\newblock Tao: A large-scale benchmark for tracking any object.
\newblock In {\em European conference on computer vision}, pages 436--454. Springer, 2020.

\bibitem{ESS2019}
Riham Elhabyan, Wei Shi, and Marc St-Hilaire.
\newblock Coverage protocols for wireless sensor networks: Review and future directions.
\newblock {\em Journal of Communications and Networks}, 21(1):45--60, 2019.

\bibitem{elhabyan2019coverage}
Riham Elhabyan, Wei Shi, and Marc St-Hilaire.
\newblock Coverage protocols for wireless sensor networks: Review and future directions.
\newblock {\em Journal of Communications and Networks}, 21(1):45--60, 2019.

\bibitem{fabbri2021motsynth}
Matteo Fabbri, Guillem Bras{\'o}, Gianluca Maugeri, Orcun Cetintas, Riccardo Gasparini, Aljo{\v{s}}a O{\v{s}}ep, Simone Calderara, Laura Leal-Taix{\'e}, and Rita Cucchiara.
\newblock Motsynth: How can synthetic data help pedestrian detection and tracking?
\newblock In {\em Proceedings of the IEEE/CVF International Conference on Computer Vision}, pages 10849--10859, 2021.

\bibitem{geiger2012we}
Andreas Geiger, Philip Lenz, and Raquel Urtasun.
\newblock Are we ready for autonomous driving? the kitti vision benchmark suite.
\newblock In {\em 2012 IEEE conference on computer vision and pattern recognition}, pages 3354--3361. IEEE, 2012.

\bibitem{guo2011sparsity}
Di Guo, Xiaobo Qu, Lianfen Huang, and Yan Yao.
\newblock Sparsity-based spatial interpolation in wireless sensor networks.
\newblock {\em Sensors}, 11(3):2385--2407, 2011.

\bibitem{HSGW2020}
James~Z Hare, Junnan Song, Shalabh Gupta, and Thomas~A Wettergren.
\newblock Pose. r: Prediction-based opportunistic sensing for resilient and efficient sensor networks.
\newblock {\em ACM Transactions on Sensor Networks (TOSN)}, 17(1):1--41, 2020.

\bibitem{he2016deep}
Kaiming He, Xiangyu Zhang, Shaoqing Ren, and Jian Sun.
\newblock Deep residual learning for image recognition.
\newblock In {\em Proceedings of the IEEE conference on computer vision and pattern recognition}, pages 770--778, 2016.

\bibitem{he2019bag}
Tong He, Zhi Zhang, Hang Zhang, Zhongyue Zhang, Junyuan Xie, and Mu Li.
\newblock Bag of tricks for image classification with convolutional neural networks.
\newblock In {\em Proceedings of the IEEE/CVF Conference on Computer Vision and Pattern Recognition}, pages 558--567, 2019.

\bibitem{howard2020fastai}
Jeremy Howard and Sylvain Gugger.
\newblock Fastai: a layered api for deep learning.
\newblock {\em Information}, 11(2):108, 2020.

\bibitem{Iakubovskii:2019}
Pavel Iakubovskii.
\newblock Segmentation models pytorch.
\newblock \url{https://github.com/qubvel/segmentation_models.pytorch}, 2019.

\bibitem{SqueezeNet}
Forrest~N. Iandola, Song Han, Matthew~W. Moskewicz, Khalid Ashraf, William~J. Dally, and Kurt Keutzer.
\newblock Squeezenet: Alexnet-level accuracy with 50x fewer parameters and $<$0.5mb model size.
\newblock {\em arXiv:1602.07360}, 2016.

\bibitem{johnson2017clevr}
Justin Johnson, Bharath Hariharan, Laurens Van Der~Maaten, Li Fei-Fei, C Lawrence~Zitnick, and Ross Girshick.
\newblock Clevr: A diagnostic dataset for compositional language and elementary visual reasoning.
\newblock In {\em Proceedings of the IEEE conference on computer vision and pattern recognition}, pages 2901--2910, 2017.

\bibitem{kingma2014adam}
Diederik~P Kingma and Jimmy Ba.
\newblock Adam: A method for stochastic optimization.
\newblock {\em arXiv preprint arXiv:1412.6980}, 2014.

\bibitem{kingma2013auto}
Diederik~P Kingma and Max Welling.
\newblock Auto-encoding variational bayes.
\newblock {\em arXiv preprint arXiv:1312.6114}, 2013.

\bibitem{koh2021wilds}
Pang~Wei Koh, Shiori Sagawa, Henrik Marklund, Sang~Michael Xie, Marvin Zhang, Akshay Balsubramani, Weihua Hu, Michihiro Yasunaga, Richard~Lanas Phillips, Irena Gao, et~al.
\newblock Wilds: A benchmark of in-the-wild distribution shifts.
\newblock In {\em International Conference on Machine Learning}, pages 5637--5664. PMLR, 2021.

\bibitem{krahenbuhl2018free}
Philipp Kr{\"a}henb{\"u}hl.
\newblock Free supervision from video games.
\newblock In {\em Proceedings of the IEEE conference on computer vision and pattern recognition}, pages 2955--2964, 2018.

\bibitem{krizhevsky2009learning}
Alex Krizhevsky, Geoffrey Hinton, et~al.
\newblock Learning multiple layers of features from tiny images.
\newblock 2009.

\bibitem{larochelle2007empirical}
Hugo Larochelle, Dumitru Erhan, Aaron Courville, James Bergstra, and Yoshua Bengio.
\newblock An empirical evaluation of deep architectures on problems with many factors of variation.
\newblock In {\em Proceedings of the 24th international conference on Machine learning}, pages 473--480, 2007.

\bibitem{lecun1998mnist}
Yann LeCun.
\newblock The mnist database of handwritten digits.
\newblock {\em http://yann. lecun. com/exdb/mnist/}, 1998.

\bibitem{li2022geoai}
Wenwen Li and Chia-Yu Hsu.
\newblock Geoai for large-scale image analysis and machine vision: Recent progress of artificial intelligence in geography.
\newblock {\em ISPRS International Journal of Geo-Information}, 11(7):385, 2022.

\bibitem{lin2014microsoft}
Tsung-Yi Lin, Michael Maire, Serge Belongie, James Hays, Pietro Perona, Deva Ramanan, Piotr Doll{\'a}r, and C~Lawrence Zitnick.
\newblock Microsoft coco: Common objects in context.
\newblock In {\em European conference on computer vision}, pages 740--755. Springer, 2014.

\bibitem{madry2017towards}
Aleksander Madry, Aleksandar Makelov, Ludwig Schmidt, Dimitris Tsipras, and Adrian Vladu.
\newblock Towards deep learning models resistant to adversarial attacks.
\newblock {\em arXiv preprint arXiv:1706.06083}, 2017.

\bibitem{masjedi2020prediction}
Ali Masjedi and Melba~M Crawford.
\newblock Prediction of sorghum biomass using time series uav-based hyperspectral and lidar data.
\newblock In {\em IGARSS 2020-2020 IEEE International Geoscience and Remote Sensing Symposium}, pages 3912--3915. IEEE, 2020.

\bibitem{plott2010mental}
Sean Plott.
\newblock Starcraft ii mental checklist, 2011.

\bibitem{quinonero2009dataset}
Joaquin Qui{\~n}onero-Candela, Masashi Sugiyama, Neil~D Lawrence, and Anton Schwaighofer.
\newblock {\em Dataset shift in machine learning}.
\newblock Mit Press, 2009.

\bibitem{robinson2019large}
Caleb Robinson, Le Hou, Kolya Malkin, Rachel Soobitsky, Jacob Czawlytko, Bistra Dilkina, and Nebojsa Jojic.
\newblock Large scale high-resolution land cover mapping with multi-resolution data.
\newblock In {\em Proceedings of the IEEE Conference on Computer Vision and Pattern Recognition}, pages 12726--12735, 2019.

\bibitem{ronneberger2015u}
Olaf Ronneberger, Philipp Fischer, and Thomas Brox.
\newblock U-net: Convolutional networks for biomedical image segmentation.
\newblock In {\em International Conference on Medical image computing and computer-assisted intervention}, pages 234--241. Springer, 2015.

\bibitem{su2019one}
Jiawei Su, Danilo~Vasconcellos Vargas, and Kouichi Sakurai.
\newblock One pixel attack for fooling deep neural networks.
\newblock {\em IEEE Transactions on Evolutionary Computation}, 23(5):828--841, 2019.

\bibitem{sun2020scalability}
Pei Sun, Henrik Kretzschmar, Xerxes Dotiwalla, Aurelien Chouard, Vijaysai Patnaik, Paul Tsui, James Guo, Yin Zhou, Yuning Chai, Benjamin Caine, et~al.
\newblock Scalability in perception for autonomous driving: Waymo open dataset.
\newblock In {\em Proceedings of the IEEE/CVF conference on computer vision and pattern recognition}, pages 2446--2454, 2020.

\bibitem{vinyals2017starcraft}
Oriol Vinyals, Timo Ewalds, Sergey Bartunov, Petko Georgiev, Alexander~Sasha Vezhnevets, Michelle Yeo, Alireza Makhzani, Heinrich K{\"u}ttler, John Agapiou, Julian Schrittwieser, et~al.
\newblock Starcraft ii: A new challenge for reinforcement learning.
\newblock {\em arXiv preprint arXiv:1708.04782}, 2017.

\bibitem{wang2020robust}
Angtian Wang, Yihong Sun, Adam Kortylewski, and Alan~L Yuille.
\newblock Robust object detection under occlusion with context-aware compositionalnets.
\newblock In {\em Proceedings of the IEEE/CVF Conference on Computer Vision and Pattern Recognition}, pages 12645--12654, 2020.

\bibitem{wang2019learning}
Qi Wang, Junyu Gao, Wei Lin, and Yuan Yuan.
\newblock Learning from synthetic data for crowd counting in the wild.
\newblock In {\em Proceedings of the IEEE/CVF conference on computer vision and pattern recognition}, pages 8198--8207, 2019.

\bibitem{Xia_2018_CVPR}
Gui-Song Xia, Xiang Bai, Jian Ding, Zhen Zhu, Serge Belongie, Jiebo Luo, Mihai Datcu, Marcello Pelillo, and Liangpei Zhang.
\newblock Dota: A large-scale dataset for object detection in aerial images.
\newblock In {\em The IEEE Conference on Computer Vision and Pattern Recognition (CVPR)}, June 2018.

\bibitem{xie2021segformer}
Enze Xie, Wenhai Wang, Zhiding Yu, Anima Anandkumar, Jose~M Alvarez, and Ping Luo.
\newblock Segformer: Simple and efficient design for semantic segmentation with transformers.
\newblock {\em Advances in Neural Information Processing Systems}, 34:12077--12090, 2021.

\bibitem{xie2017aggregated}
Saining Xie, Ross Girshick, Piotr Doll{\'a}r, Zhuowen Tu, and Kaiming He.
\newblock Aggregated residual transformations for deep neural networks.
\newblock In {\em Proceedings of the IEEE conference on computer vision and pattern recognition}, pages 1492--1500, 2017.

\bibitem{yan2022lawin}
Haotian Yan, Chuang Zhang, and Ming Wu.
\newblock Lawin transformer: Improving semantic segmentation transformer with multi-scale representations via large window attention.
\newblock {\em arXiv preprint arXiv:2201.01615}, 2022.

\bibitem{younis2014topology}
Mohamed Younis, Izzet~F Senturk, Kemal Akkaya, Sookyoung Lee, and Fatih Senel.
\newblock Topology management techniques for tolerating node failures in wireless sensor networks: A survey.
\newblock {\em Computer networks}, 58:254--283, 2014.

\bibitem{zheng2016person}
Liang Zheng, Yi Yang, and Alexander~G Hauptmann.
\newblock Person re-identification: Past, present and future.
\newblock {\em arXiv preprint arXiv:1610.02984}, 2016.

\bibitem{zhou2017places}
Bolei Zhou, Agata Lapedriza, Aditya Khosla, Aude Oliva, and Antonio Torralba.
\newblock Places: A 10 million image database for scene recognition.
\newblock {\em IEEE Transactions on Pattern Analysis and Machine Intelligence}, 2017.

\bibitem{zhou2022domain}
Kaiyang Zhou, Ziwei Liu, Yu Qiao, Tao Xiang, and Chen~Change Loy.
\newblock Domain generalization: A survey.
\newblock {\em IEEE Transactions on Pattern Analysis and Machine Intelligence}, 2022.

\end{thebibliography}
}

\newpage
\appendix

\begin{table*}[!ht]
    \centering
    \caption{Dataset Nutrition Label for the \datasetname dataset.}
    \includegraphics[width=\textwidth]{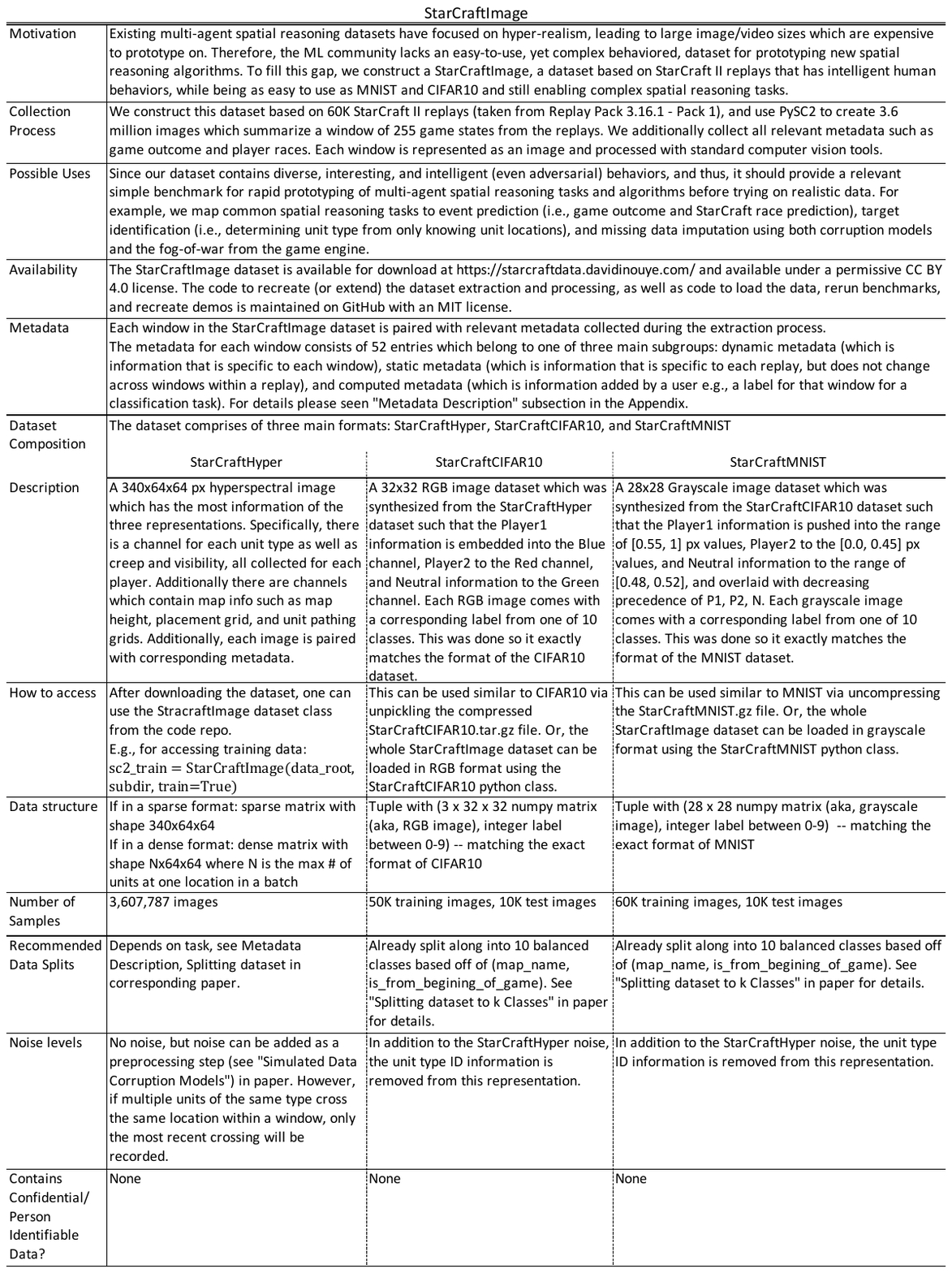}
    \label{tab:Dataset-Nutrition-Label}
\end{table*}
\clearpage

\section{Dataset Availability, Licensing, and Management}
The \datasetname dataset is available for download at \dataseturl which contains the full extracted data from the 3.6 million windows, the metadata for all windows, the StarCraftMNIST train/test datasets, and the StarCraftCIFAR10 train/test datasets. 
The code to recreate (or extend) the dataset extraction and processing, as well as code to load the data, rerun benchmarks, and recreate demos, can be found at the previous link, and is maintained on GitHub.
Instructions for loading and using the dataset can be found in the README in the dataset as well as in the code repository.
The dataset has been openly published under a permissive CC BY 4.0 license, and the code has been openly published on GitHub with an MIT license.
The authors bear all responsibility in case of violation of rights and confirm the CC license for the provided datasets.

\section{Direct Loading of Window Data}
While we encourage using the corresponding PyTorch dataset classes that we have developed (one class for each representation) to load in \datasetname data, one can also directly access the data by loading in the relevant \texttt{.png} file and metadata row for each window.
To assist with this direct data access, we now describe the data structure used to store the image data for each window (\ie how to correspond each \texttt{.png} to the hyperspectral format $H$ discussed in \autoref{ssec:sensorhyper}).

\begin{figure}[!ht]

    \begin{subfigure}[b]{\linewidth}
    \includegraphics[width=\linewidth]{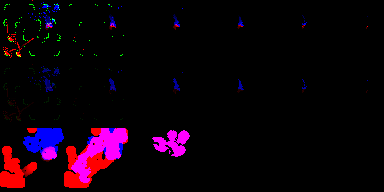}
    \caption{Max number of units overlapped is 6.}
    \end{subfigure}
    \begin{subfigure}[b]{\linewidth}
    \includegraphics[width=0.666667\linewidth]{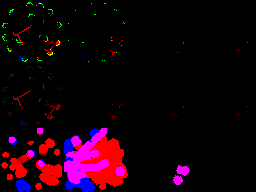}
    \caption{Max number of units overlapped is 4.}
    \end{subfigure}
    \begin{subfigure}[b]{\linewidth}
    \includegraphics[width=0.666667\linewidth]{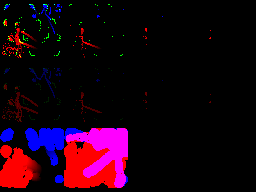}
    \caption{Max number of units overlapped is 4.}
    \end{subfigure}
    \caption{Three examples of the dense bag-of-units \texttt{.png} show how the hyperspectral image data for a window is stored in a simple \texttt{.png} file.
    The hyperspectral information is represented by tiling 64 x 64 RGB images.
    The first row is the unit timestamps (0-255), the second row is the unit ids (0-255), and the third row contains map state information (\texttt{is\_visible}, \texttt{is\_seen}, and \texttt{creep}).
    The blue channel encodes player 1, the red channel encodes player 2, and the green channel encodes neutral elements.
    We note how the width of the image varies, as it is determined by how many overlapping units at the same location there are in that window. For example, the top example had 6 units overlapping at one location, so it has a width of six 64x64 images whereas the other two only had a max of 4 units overlapping at one location.}
    \label{fig:bag-image_examples}
    \vspace{-1 em}
\end{figure}

As a reminder, the bag-of-units representation collapses the channel, axis of our hyperspectral image $H$ into $k$ ID matrices and $k$ timestamp matrices of size $(64, 64)$, where the ID matrix contains the $PID$ of the units present at each $(x,y)$ coordinate, the timestamp matrices contain the corresponding timestamp that the unit was last seen, and $k$ is the max number of units present at one $(x,y)$ location in $H$, seen in the top right of \autoref{fig:hyperspectral-overview}.
We can further compress this bag-of-units representation by stacking the bag-of-units for player 1, player 2, and neutral to match the RGB structure of a \texttt{.png} image, where the red channel corresponds to player 2, the blue channel to player 1, and the green channel corresponds to the neutral units (e.g., mineral deposits).
To fit the structure of a $RGB$ image, we can tile the bag-of-units into rows where the first $RGB$ row corresponds to the timestamp matrices and the second row corresponds to the $PID$ matrices.
Finally, we add a third row to record the map state information for each player, specifically, the map state row contains: [RGB 'is\_visible', RGB 'is\_seen', RGB 'creep'].
This leaves us with a $RGB$ \texttt{.png} image with height $3*64$ and width $k*64$.
Examples of this can be seen in \autoref{fig:bag-image_examples}.

\section{Broader Impact}
We introduced this spatial reasoning dataset to allow for quick prototyping of complex multi-agent spatial reasoning ML models and easy benchmarking to compare models (similar to the use cases of MNIST and CIFAR10).
While our dataset contains complex dynamics that are based on real human actions, it is still a simulation-based dataset, and thus methods tested on this dataset should be further tested in real-world cases before a real-world deployment.
Additionally, since this dataset is a general spatial reasoning dataset that can either be directly applied or easily adapted to real-world cases, there is an opportunity for this dataset to be used for tasks that have a negative societal impact (e.g., unauthorized surveillance/tracking). 
We do not condone the usage of this dataset for the development of harmful models for such negative tasks.
Furthermore, since our replays are created by humans and have personal metadata like the actions per minute (APM) and match-making rating (MMR) for each player, this could possibly be used to uniquely attribute a replay to a player.
However, this likely is only possible for extreme APM, MMR values (\eg the top MMR value), and even then, APM is match-specific and a player's MMR updates with each match. 
Finally, all replays were freely uploaded in an open-source manner and (to the best of our knowledge) contain no personally identifying information (\eg name of the uploader, upload IP address, etc.).

\section{Metadata Description and Suggested Classwise Splits}
\label{sec:metadata-description}
In this section we discuss the metadata collected alongside the image data for each window in \datasetname.
\begin{figure*}
    \centering
    \includegraphics[width=0.8\textwidth]{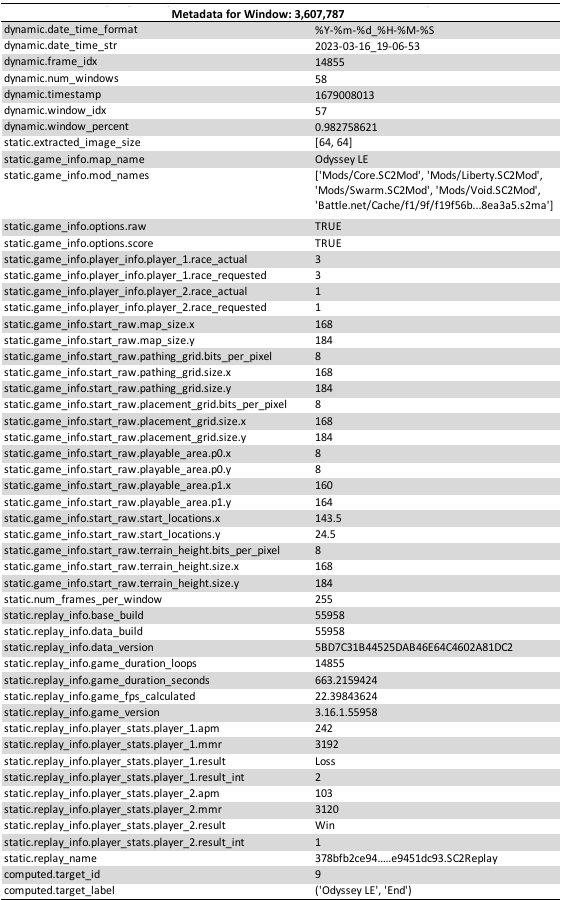}
    \caption{An example of the metadata collected for each window of a replay. 
    Descriptions of the key, value pairs are given in \autoref{sec:metadata-description}.}
    \label{fig:metadata-row-example}
\end{figure*}

\subsection{Metadata Description}
\begin{figure*}[ht!]
    \centering
    \includegraphics[width=\textwidth]{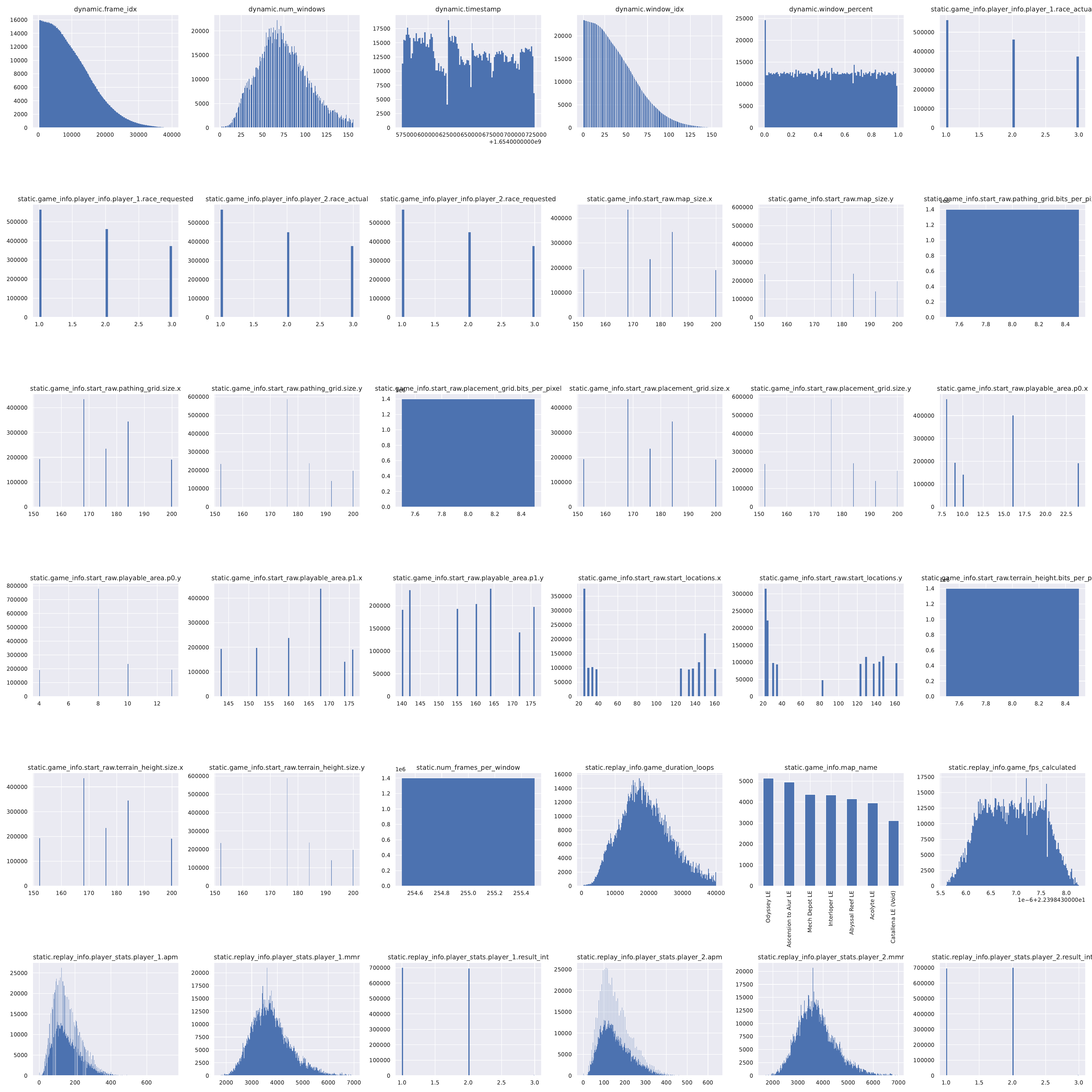}
    \caption{Histograms of the numerical values entries in the metadata (best viewed zoom in).
    To keep the x-axis interpretable, we excluded any outliers which have a game\_duration\_loop $>$ 40,000 and any replays which hold a negative MMR value for either player (which is likely a result of a bug in the \texttt{.SC2Replay} file). 
    Any histograms which are fully rectangular are static values at the center point of the bin (\eg \texttt{static.num\_frames\_per\_window}=255 for all windows).}
    \label{fig:metadata-histograms}
\end{figure*}

For each window in \datasetname, we also collected relevant match/window metadata, which can be seen in \autoref{fig:metadata-row-example}.
Each entry belongs to one of three main subgroups: dynamic metadata (which is information that is specific to each window), static metadata (which is information that is specific to each replay but does not change across windows within a replay), and computed metadata (which is information added by a user e.g., a label for that window for a classification task).
Namely, the dynamic metadata contains a vector of tabular features for both player1 and player2 such as resource counts for each player.
Specfically, these tabular features correspond to: \texttt{['player\_id', 'minerals', 'vespene', 
                      'food\_used', 'food\_cap', 'food\_army', 'food\_workers', 
                      'idle\_worker\_count', 'army\_count', 'warp\_gate\_count', 
                      'larva\_count']}.
Additionally, the dynamic metadata contains: \texttt{date\_time\_str} which is a string representing the date that window was added to the dataset, \texttt{frame\_idx} which is the frame index within a replay which corresponds to the last frame included in a window (e.g., if a window's \texttt{dynamic.frame\_idx}=1000 then that window summarizes frames 745 to 1000 of the given replay).
The dynamic \texttt{window\_percent} corresponds to how far into a match that window takes place, represented as a fraction.
For the \texttt{static} metadata, this is broken into \texttt{game\_info} (which corresponds to information that is mostly match specific such as map information) and \texttt{replay\_info} (which replays to information about the replay file and the players contained in the file).
In the \texttt{game\_info}, the race information is encoded following the PySC2 convention where Terran $=1$, Zerg $=2$, Protoss $=3$, and Random $=4$.
The player-level information can be found in the \texttt{replay\_info.player\_stats} section where APM corresponds to the player's Actions Per Minute for that match and the player's MMR is the player's Match Making Rating (which can be thought of as a skill-level determined by Blizzard, where higher is more skilled).
We include these metadata to give more details about each window, but most importantly to allow a user to split the \datasetname dataset along these features for a specific task.
For example, if one is developing a model which should generalize to new environments, a user can split this dataset on the \texttt{static.game\_info.map\_name} feature, and use windows from five of the seven maps for training/validation and test on windows from the  remaining two maps.
To aid in determining filtering methods, histograms for numeric entries within the metadata can be seen in \autoref{fig:metadata-histograms}.

\begin{table*}[h!]
    \centering
    \caption{A per-window frequency table for the non-neutral units across all windows in the \datasetname dataset, where Avg. Per Win. corresponds to the average number of times that unit is present per window, Perc. is the number of times that unit appeared divided by the total unit appearances, and Cum Perc. is the cumulative percentage up to that row.
    Note, this analysis was performed on the 30k replay subset but should be quite similar to the 60k frequencies.} 
    \vspace{-5 em}
    \includegraphics[width=\textwidth]{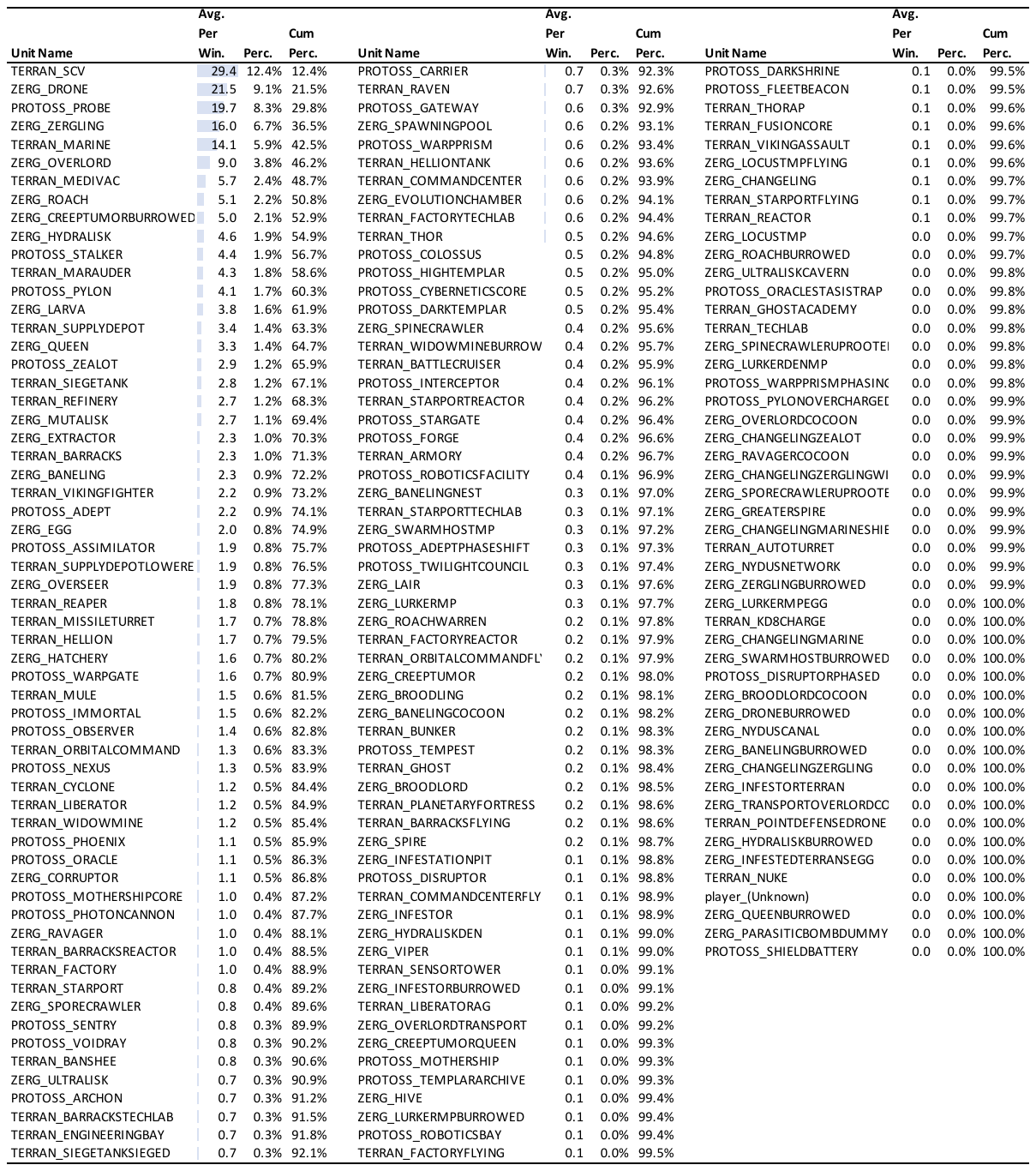}
    \label{fig:unit-counts}
\end{table*}

\subsection{Splitting dataset to k Classes}
\label{ssec:class-splitting}
When working with global-spatial reasoning tasks (\eg whole-image classification), the question of how to split this dataset into $k$ classes arises.
Thus, we suggest some possible ways to split the dataset along with simple benchmark accuracy values for comparing the difficulty of the splits for two of the most common classification schemes ML: binary classification ($k$=2) and 10-way classification ($k$=10).
For all splitting experiments we mention running below, we use the same ConvNet architecture of two convolutional layers with max-pooling in-between, three fully connected layers, all with ReLU activations, and train for 20 epochs using SGD with a learning rate of 0.001 and momentum of 0.9.

For binary classification, the obvious choice is to perform match outcome prediction (\ie ``did Player 1 win?''), as mentioned in the main body of this work.
While an important task, this can be difficult even for human experts watching a StarCraftII E-sports event as well as difficult for an AI to solve/
For example, the best model in \cite{vinyals2017starcraft} can only achieve 65\% outcome prediction training accuracy for frames taken 15 minutes into a game, and when tested on the grayscale StarCraftMNIST and RGB StarCraftCIFAR10 datasets split on the match outcome variable  (which include windows throughout all points in the game rather than just mid-to-end game), we report 57.9\% and 59.4\% test accuracy, respectively, on the same task. 
A binary prediction task that is more easily interpreted is the task of predicting if a window comes from the first half or second half of a replay (``is \texttt{dynamic.window\_percent} < 0.5?'').
This is also somewhat easier to solve (we report a testing accuracy of 74.3\% and 76.9\% for grayscale StarCraftMNIST and RGB StarCraftCIFAR10 datasets with this split, respectively), while still requiring a model to learn environmental dynamics to solve.

Splitting the \datasetname dataset for 10-way classification (\eg StarCraftMNIST and StarCraftCIFAR10) is more difficult since there are no natural way to split the dataset 10 ways.
The splitting method which most closely aligns with spatial reasoning problems is likely splitting via player race information + match outcome prediction, as this requires learning battlefield strategy/dynamics (for outcome prediction) and understanding of unit information (for player race prediction).
Thus, this is the 10-way splitting method suggested in \autoref{ssec:metadata-analysis} in the main body of this work.
However, from a purely ML perspective, this is an extremely difficult classification problem; which is supported by our testing accuracy of 26.4\% and 28\% for StarCraftMNIST and StarCraftCIFAR10 datasets created with the split detailed in \autoref{ssec:metadata-analysis}.
Thus, for purposes with a stronger abstract ML focus, we suggest a 10-way split that combines the ``is\_beginning\_or\_end'' binary variable from above with a prediction of the map\_name. 
This task requires the model to learn environment information for the map\_name and battlefield dynamics for the beginning/end prediction and is more solvable than a split requiring match outcome prediction.
Specifically, since there are 7 maps in total, we suggest subsampling to only 5 maps, then further splitting each of these 5 map groups into ``beginning `` and ``end'' groups (based on whether or not the window takes place in the beginning 50\% of the match), to get 10 classes.
Examples from such a split can be seen in figure \autoref{fig:map-ten-class}, and in our experiments, we received a testing accuracy of 77.2\% and 77.9\% for StarCraftMNIST and StarCraftCIFAR10 datasets, respectively.
Heuristically, we have found this 10-way split to be a good balance between problem realism and difficulty/human interpretability, and thus we will be using it for the following task demonstrations.

\section{Benchmark Evaluations On Multi-Agent Spatial Reasoning Tasks}
\label{sec:benchmarks-on-sn-tasks-APPENDIX}
In this section, we report benchmark results on 4 benchmark multi-agent spatial reasoning  tasks, which incorporate training 60 U-Net-based \cite{ronneberger2015u} models. 
Unlike the benchmark results in the main paper, (e.g., \autoref{tab:benchmark-results-MAIN}), the results seen here and throughout the rest of the appendix are trained on a smaller StarCraftImage dataset (specifically, these results are generated from a random 1.8 million window subset, i.e. a random 50\% subset of the main dataset).
This was done to allow for faster model training, thus allowing us to add more models beyond the three ResNet models seen in the main paper (specifically, 60 models were trained on this smaller dataset).

\begin{figure}[!ht]
    \includegraphics[width=\columnwidth, trim=0 0 0 6.18cm, clip]{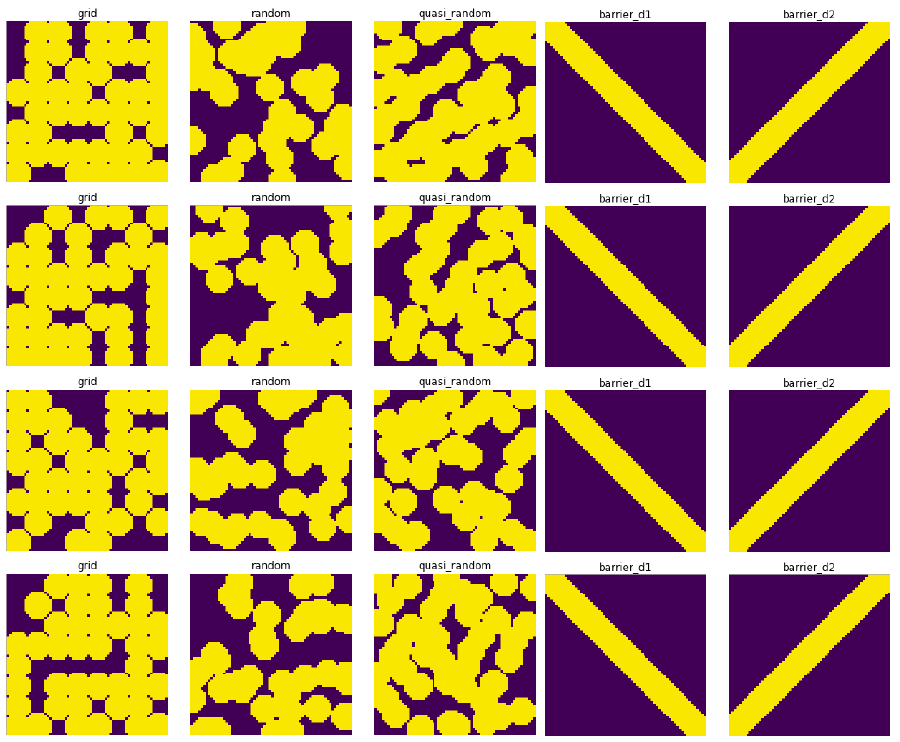}
    \caption{
    Example masks for sensor network simulations containing 50 sensors with five different placement strategies [grid, random, quasi-random, barrier-d1, barrier-d2] (where yellow is a location that is visible). 
    Each sensor has a radius of 5.5. pixels and a 20\% chance of being dropped during the sensor fusion process (e.g., due to a communication failure).
    }
    \label{fig:sensor-simulation-masks}
\end{figure}

The four benchmark tasks consist of two tasks on unit type identification and two tasks for unit tracking (next hyperspectral window prediction).
Both task sets consist of first training and evaluating on ``clean'' (unaltered) data as well as a second task of training on data which is first passed through a simulation of a noisy sensor network.
This simulation consists of 50 sensors with a radius of 5.5 pixels with different sensor placement methodologies (e.g., grid, random, quasi-random, and diagonal barrier) and communication failures during sensor fusion. 
For reference, grid-based placements are commonly used in environmental monitoring data sets where they are optimally placing sensors to cover the environment space \cite{li2022geoai}.
Random and Quasi-random deployments are typical when sensors cannot be placed optimally (e.g., when they are dropped out of planes/helicopters).
The barrier placements come from the well-studied barrier coverage problem, which is commonly used for border surveillance, road monitoring, etc. 
For further examples studying these different coverage types see \cite{elhabyan2019coverage} which provides a taxonomy for different coverage protocols including the ones mentioned above.
For a study on different failure types, including the ones seen here, see \cite{younis2014topology}.
Examples of the different placement types can be seen in \autoref{fig:sensor-simulation-masks}.

 \subsection{Unit Type Identification}
As discussed in \autoref{sec:ml-tasks-on-sn}, the goal of this task is to train a model to take a 64 x 64 RGB image (similar to the format of the StarCraftCIFAR10 images) as input and to output a 64 x 64 matrix corresponding to the unit ids for each location.
This problem is analogous to fine-grained multi-object detection, where given raw images (\eg satellite images), our goal is to predict what kind of unit is present at each location (if there is a unit present at all).
For example, for a given window, if there is a non-zero value at location, $($Red$, i, j)$, then we know an enemy unit passed through location $(i, j)$ -- our goal now is to figure out that unit's type (e.g., ZERG\_QUEEN, PROTOSS\_ORACLE, etc.).

\begin{figure*}[!ht]
    \centering
    \includegraphics[width=\textwidth]{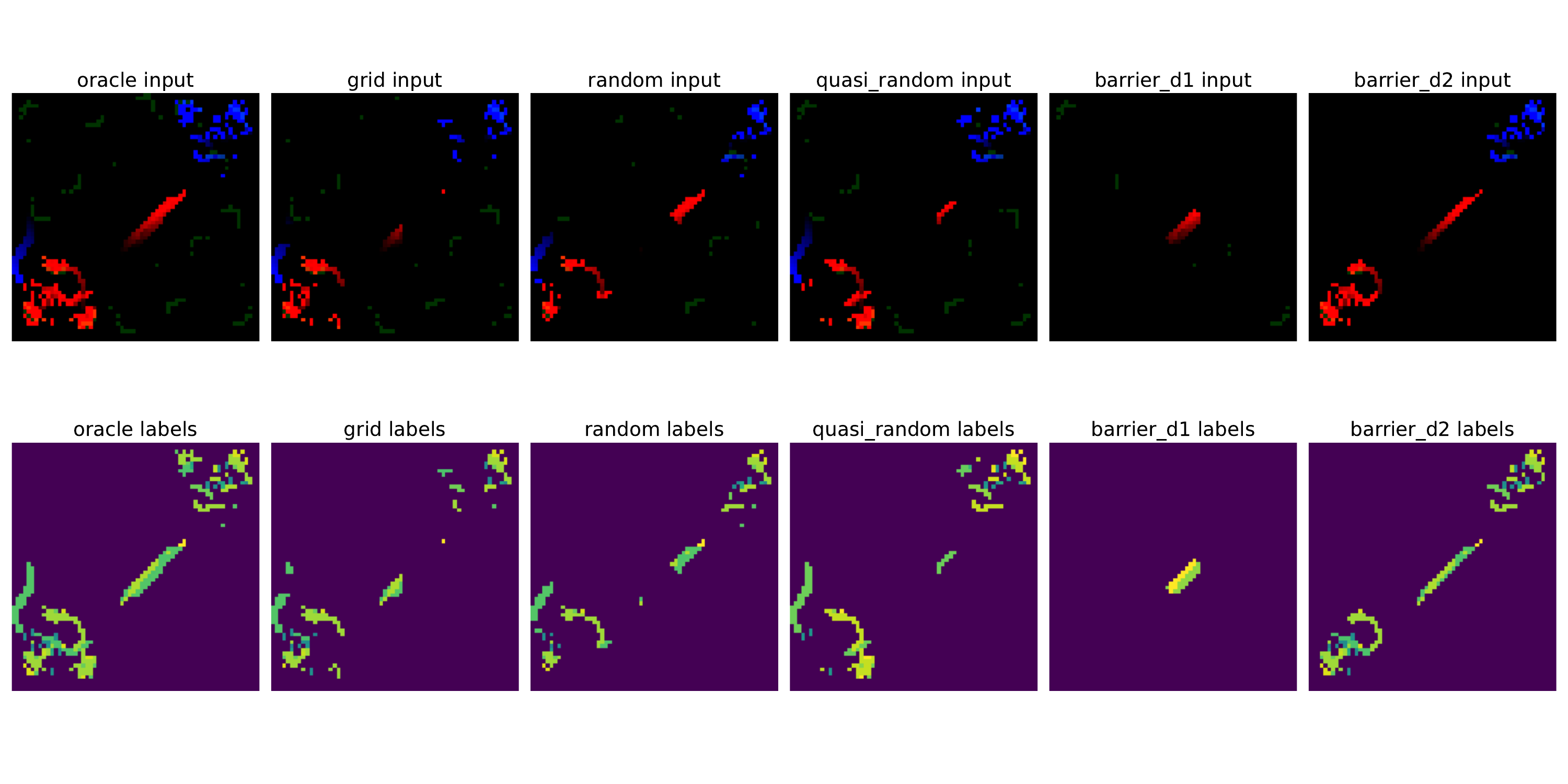}
    \caption{
    Example input-output pairs for the sample from the Unit Type Identification task with the clean representation as well as representations which are prepossessed by corruption simulations via faulty sensing networks five different placement strategies [grid, random, quasi-random, barrier-d1, barrier-d2] (where yellow is a location that is visible). 
    Note that during training, the labels are also masked to simulate training on noisily labeled data.
    }
    \label{fig:unit-id-demo-plots}
\end{figure*}

We used the unaltered RGB images as input and synthesized the 64 x 64 unit id label matrix from the StarCraftHyper dataset.
In cases where multiple units were present at the same location (which is possible since a window covers a span of 255 in-game frames/seconds), we set the id for that location to that of the most recent unit.
We used the FastAI library \cite{howard2020fastai} to train six U-Net models \cite{ronneberger2015u} with backbones: ResNet18, ResNet34 \cite{he2016deep}, Squeezenet1\_0, Squeezenet1\_1 \cite{SqueezeNet}, XResNet18, XResNet34 \cite{he2019bag}, on this dataset for 10 epochs with cross-entropy loss, a batch size of (512 for ResNet*, 256 for Squeezenet*, and 512 for XResNet*), and default FastAI configuration settings.
We trained each of the above models on the clean (\ie noiseless) dataset as well as on all five sensor placement variations where both the input images and output label matrices were masked by the generated sensor masks (see \autoref{fig:unit-id-demo-plots} for examples), yielding 36 models in total.
During testing, we tested all models on held-out \emph{clean} data, which simulates the situation where one has noisy training data but wants to evaluate an algorithm with respect to clean ground truth data for final evaluation.
We report the Cross-Entropy error, Unit Accuracy (was the unit type correctly predicted), and the averaged Dice coefficient for all models in \autoref{tab:unit-identification-results}.
As expected, there is significant performance derogation across models when moving from the clean data to the noisy data, and this is most evident in the diagonal barrier placements.
As seen by the unit accuracy metric, this is a hard problem (there are 340 possible unit ids for each location), which we hope future work will be able to innovate upon.

\begin{table*}[!ht]
    \centering
    \caption{Results for the Unit Identification Benchmarks.}
    \label{tab:unit-identification-results}
    \includegraphics[width=\textwidth]{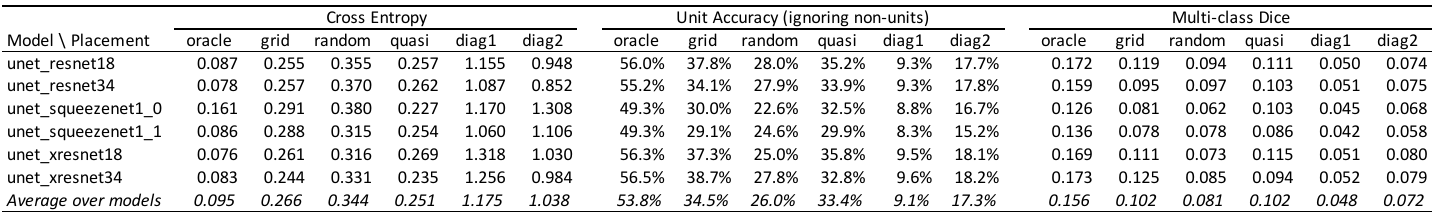}
\end{table*}

\subsection{Next Hyperspectral Window Prediction}

Here our goal is to use the StarCraftHyper dataset to train a model on the common spatial reasoning task of object tracking.
For this, we frame this task as: a given replay, we want to take the $k$\th hyperspectral window as input (with shape 340 x 64 x 64) and have a model forecast how all units will move to their locations in the $k+1$ hyperspectral window.
Specifically, the model must output the \emph{difference} (i.e., movement) between the two hyperspectral windows ($ground\_truth = (H_{k+1} - H_k)$ and has shape (340, 64, 64)).

\begin{figure*}[!ht]
    \centering
    \includegraphics[width=\textwidth]{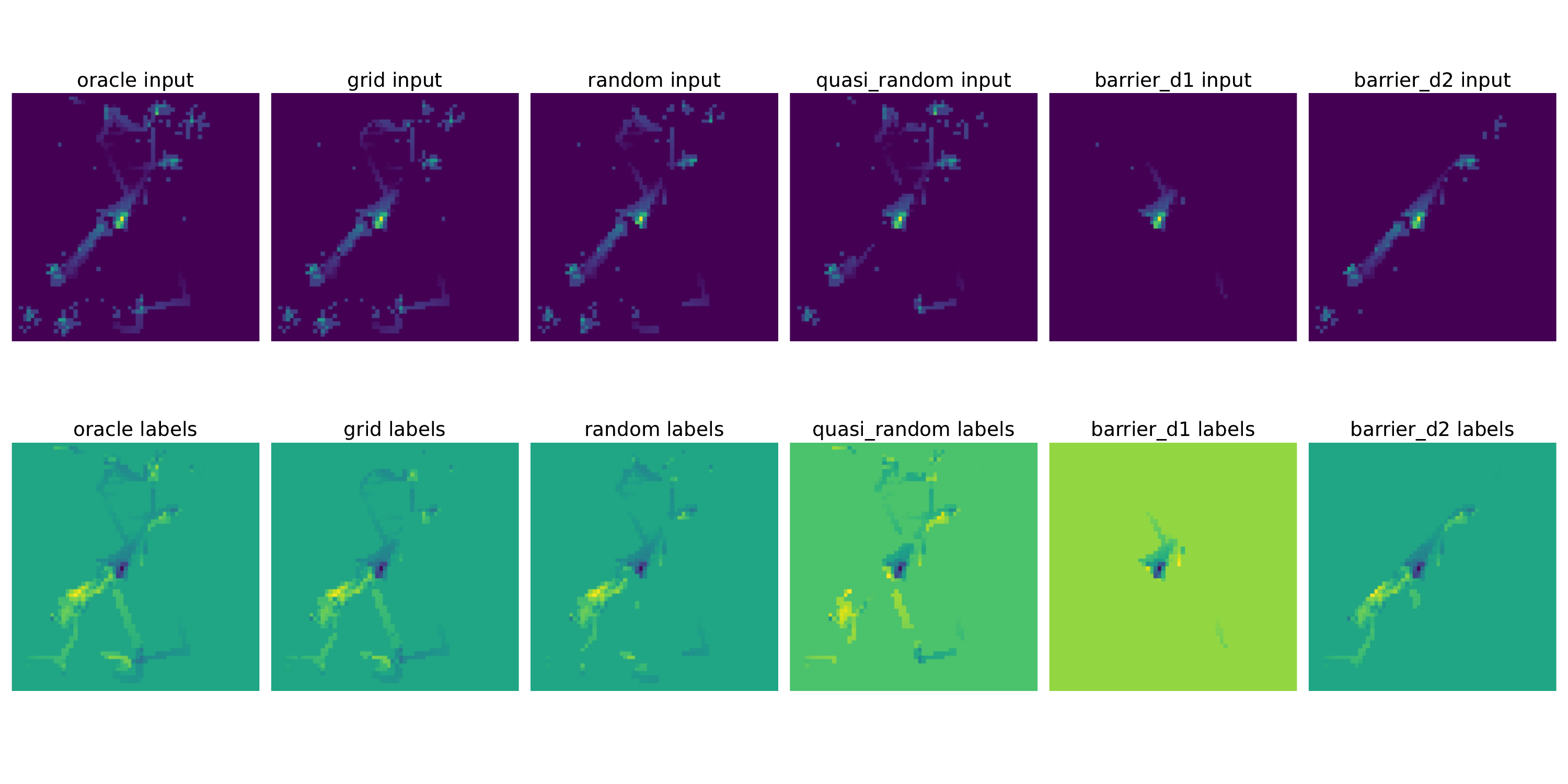}
    \caption{
    Example input output pairs of the same sample from the Next Hyperspectral Window prediction task with the clean representation as well as representations which are prepossessed by corruption simulations via faulty sensing networks with five different placement strategies [grid, random, quasi-random, barrier-d1, barrier-d2] (where yellow is a location that is visible). 
    Note that during training, the labels are also masked to simulate training on noisily labeled data.
    }
    \label{fig:next-window-demo-plots}
\end{figure*}

To do this, we use the FastAI library \cite{howard2020fastai} and the SMP library \cite{Iakubovskii:2019} to train four U-Net \cite{ronneberger2015u} models with backbones: ResNet18, ResNet34, ResNet50 \cite{he2016deep}, and ResNext50\_32x4d \cite{xie2017aggregated} on both clean versions of the dataset and five noisy versions of the dataset matching the five sensor network simulations (see \autoref{fig:next-window-demo-plots} for examples).
Due to the large size of the samples, we use a batch size of 20 across all models, and to accelerate the training process we randomly subsample the overall training data to 60K window pairs.
We train all models for 10 epochs using the Mean-Squared Error loss and otherwise default FastAI configuration settings.
We then test our models on 10K held out \emph{clean} window pairs, and report the MSE loss.
Additionally, we bin the $ground_truth$ test data into $\left[-1, 0, 1\right]$ where location $(u_{id}, i, j)$  is $-1$ if a specific unit type \emph{left} location $(i, j)$ from window $k$ to window $k+1$  (i.e. the unit moved away from that spot), $0$ means no movement happened, and $1$ means unit $u_{id}$ moved \emph{into} location $(i, j)$.
With this, we report the False Positive Rate (\% of times a model predicts movement happens when it doesn't), True Positive Rate (+) (\% of the time a model correctly predicts $+1$, a unit moved into a location), and the True Positive Rate (-).
The results for this can be seen in \autoref{tab:next-window-results}, where while the MSE is lowest for the models trained on the clean data, these models also tend to have a higher false positive rate.

\begin{table*}[!h]
    \centering
    \caption{Results for the Next Hyperspectral Window Prediction Benchmarks.}
    \label{tab:next-window-results}
    \includegraphics[width=\textwidth]{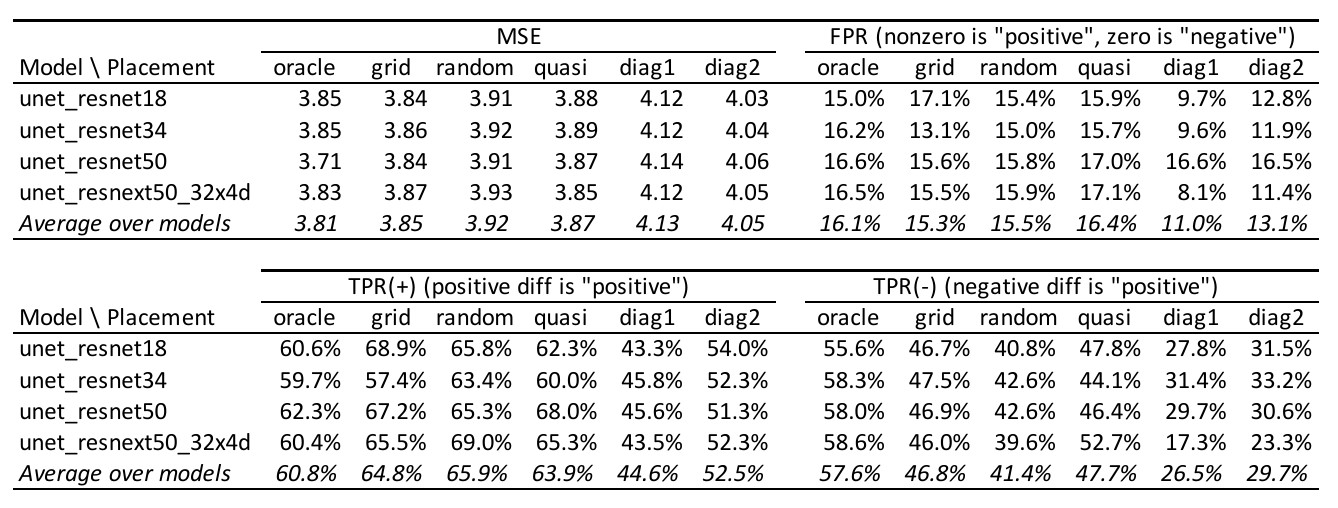}
\end{table*}

\section{Additional Demonstrations of ML tasks on the \datasetname Dataset}
\label{sec:additional-experiments-APPENDIX}
\paragraph{Predicting match outcome} As mentioned previously, this is a very difficult task which can be hard even for human experts.
\cite{vinyals2017starcraft} performed the match outcome prediction task on features extracted from PySC2, and report only 65\% outcome prediction training accuracy for frames taken as far as 15 minutes into a game.
For this task, we use two datasets: one which has the grayscale window formats of StarCraftMNIST and the other with the RGB window formats of StarCraftCIFAR10.
These datasets are constructed such that the train/test splits for the positive class (Player1IsWinner) and negative class (Player1IsNotWinner) are evenly balanced with (60k, 10k) and (50k, 10k) train, test examples for the grayscale and RGB datasets, respectively.
We found the performance of our ConvNet model (mentioned in the previous section) to be similar to that of \cite{vinyals2017starcraft}, where we see only 57.9\% test accuracy on the grayscale window dataset and 59.4\% test accuracy on the RGB window dataset.

\begin{figure*}[!ht]
    \centering
    \includegraphics[width=0.93\textwidth]{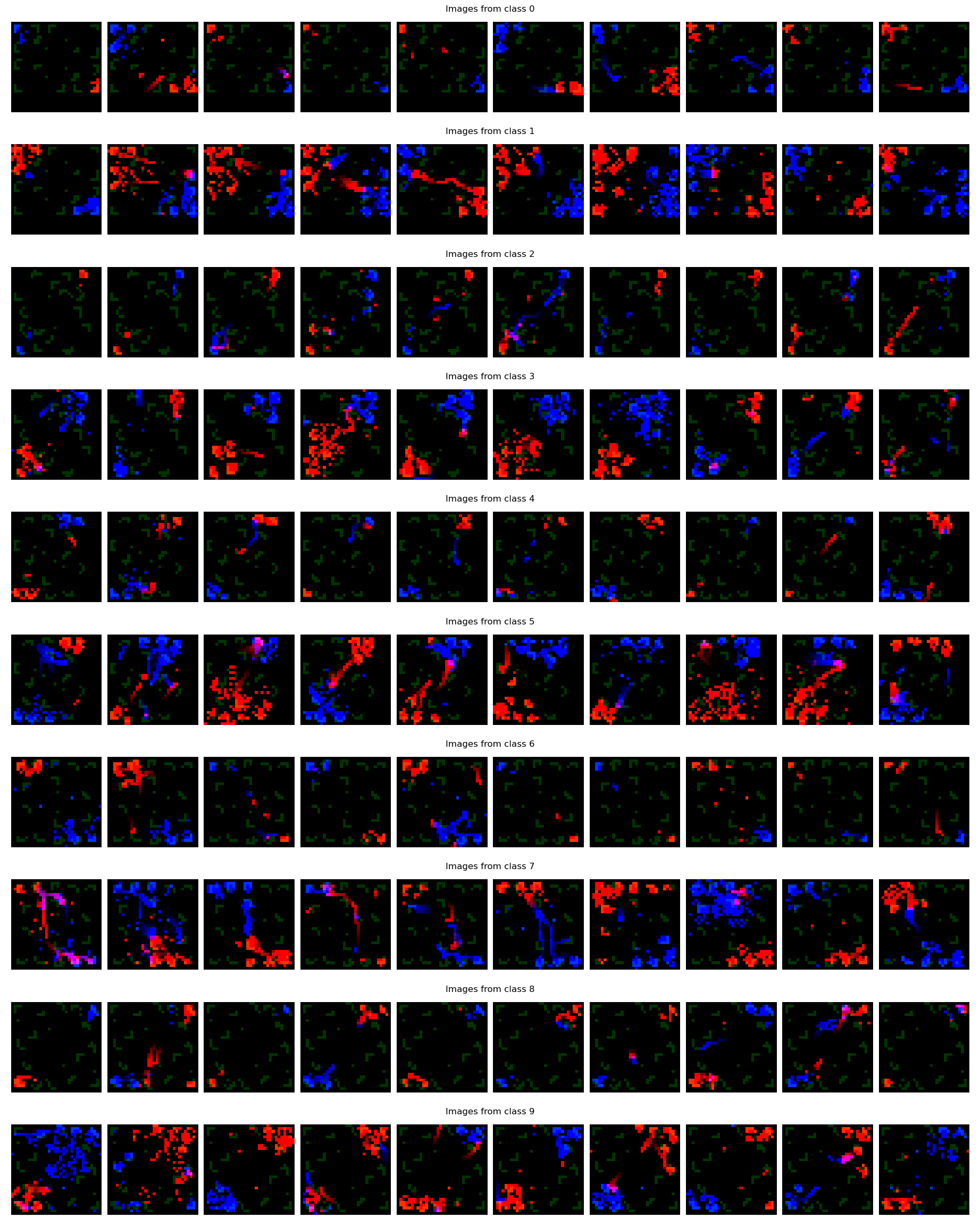}
    \vspace{2 em}
    \caption{10 random samples from each class (where each row is its own class), from the map\_name + is\_begining\_or\_end 10-way class split.
    The class label to variable information mapping is as follows: 
              Class\_0=('Acolyte LE', 'Beginning'),
              Class\_1=('Acolyte LE', 'End'),
              Class\_2=('Abyssal Reef LE', 'Beginning'),
              Class\_3=('Abyssal Reef LE', 'End'),
              Class\_4=('Ascension to Aiur LE', 'Beginning'),
              Class\_5=('Ascension to Aiur LE', 'End'),
              Class\_6=('Mech Depot LE', 'Beginning'),
              Class\_7=('Mech Depot LE', 'End'),
              Class\_8=('Odyssey LE', 'Beginning'),
              Class\_9=('Odyssey LE', 'End')).}
    \vspace{2 em}
    \label{fig:map-ten-class}
\end{figure*}

\paragraph{10 class classification} Here we use the map\_name + is\_beginning\_or\_end 10-way class split mentioned in \autoref{ssec:class-splitting} and seen in \autoref{fig:map-ten-class}.
We apply our ConvNet model to the StarCraftMNIST and StarCraftCIFAR10 datasets.
After training for 20 epochs, we received a testing accuracy of 77.2\% and 77.9\% for StarCraftMNIST and StarCraftCIFAR10, respectively.

\begin{figure}[!ht]
    \centering
    \includegraphics[height=0.65\textheight]{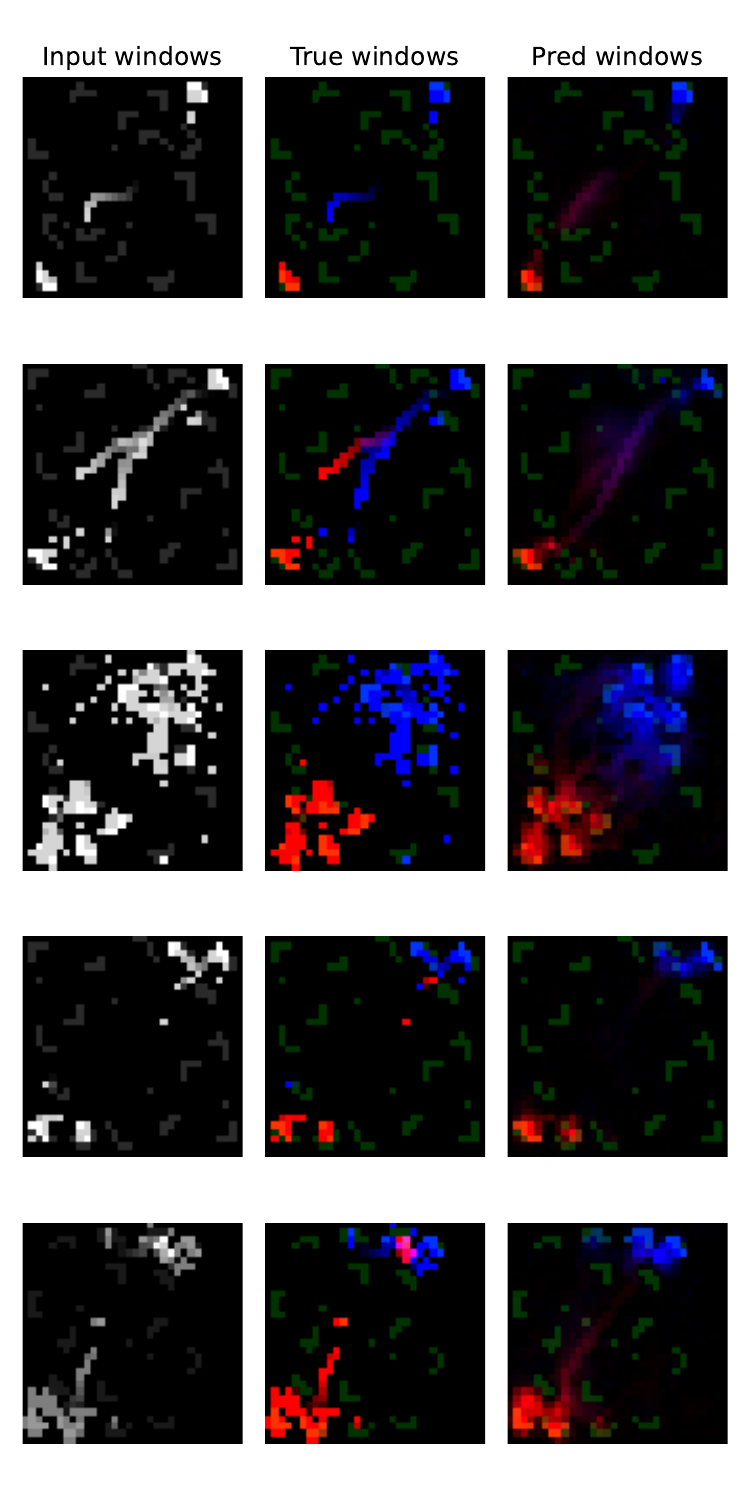}
    \caption{Five examples from the ResNet-101 model \cite{he2016deep} trained to identify the owner of each unit in a window, where this task is akin to an image colorization task where each owner (Player1, Neutral, Player2) is placed on a separate channel. As can be seen in the examples here, it is difficult at times for the model to determine the difference between Player 1 and Player 2 due to both players having random starting corners of the map.}
    \label{fig:image-colorization-demo}
\end{figure}

\paragraph{Imputing occluded objects} For this task, we simulate occluded objects via randomly masking out a circle of pixels with a radius 5px, which can be seen to approximate cloud coverage or a fused image with missing data (see left of \autoref{fig:image-colorization-demo} for examples).
For this task, our goal is to impute the missing information and to do this we implement a VAE model \cite{kingma2013auto} to denoise the inpainted image.
Our VAEImputer encoder and decoder both consist of six convolutional layers (with a 3x3 kernel, stride of 2, and padding of 1), each with batch norm and leakyReLu activations and with one fully connected layer in-between.
For the encoder, the convolutional layers consist of $[3, 32, 64, 128, 128, 512]$ channels, the decoder convolutional layers consist of $[512, 128, 128, 64, 32, 3]$ channels, and the fully connected layer takes in the 512 channels and projects this to the 64 dimensional $\mu$ and $\sigma$ latent parameter space.
We train the model for 100 epochs on StarCraftMNIST (which consists of the same 60k training images as in the 10-class map-based split), with Adam optimizer \cite{kingma2014adam} with a learning rate of 5e-3 and $\beta=(0.9, 0.999)$.
The results can be seen in \autoref{fig:image-colorization-demo}, where the model does well at imputing the missing data, at the expense of blurring the image due to artifacting from the VAE.

\begin{figure}[!ht]
    \centering
    \includegraphics[height=0.75\textheight]{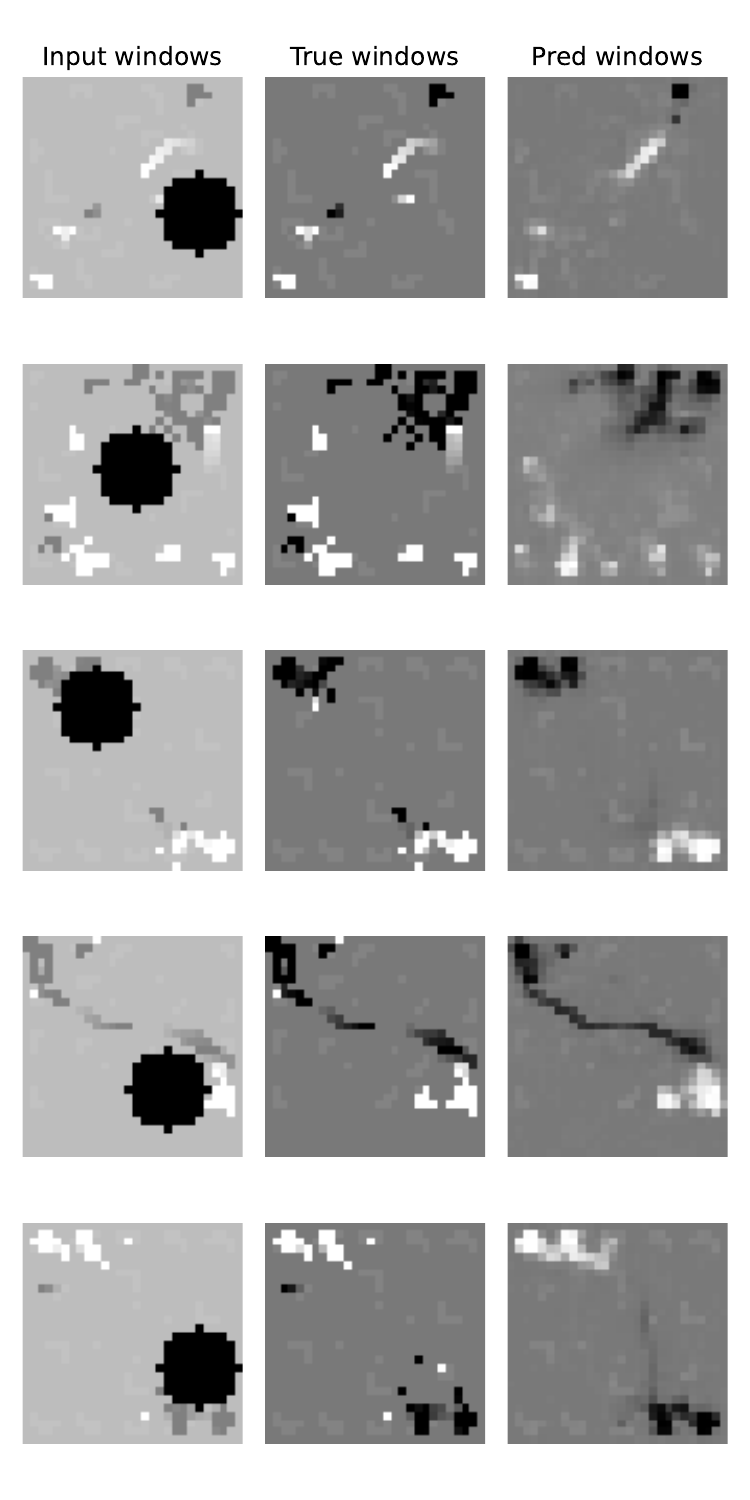}
    \caption{Five examples from the VAEImputer model trained to denoise an imputed corrupted aerial image (where in this case an occlusion with a 5px radius has been simulated). As can be seen. Note, the difference in colorization between the input windows and true windows is simply due to plotting renormalization due to the occlusion.}
    \label{fig:next-window-prediction-demo}
\end{figure}

\vspace{-1 em}
\paragraph{Unit type identification} In this task we simulate the case where a model is only given a raw image showing the presence of a unit, but not what type of unit it is. 
This can be mapped to a colorization task by first taking an RGB sample from StarCraftCIFAR10 (where each channel corresponds to a specific player's units) and averaging along the color dimension to get a grayscale image that has no owner information.
Then, to recover the owner information (i.e. whether a unit belongs to Player 1, Player 2, or Neutral), we colorize the image by predicting which channel each unit should belong to.
To do this, we use a ResNet-101 model \cite{he2016deep}, which has been adapted to have an output dimensionality of 32x32x3 (the number of pixels in a StarCraftCIFAR10 image).
We train this model on the StarCraftCIFAR10 dataset for 100 epochs using the Adam optimizer \cite{kingma2014adam} with a learning rate of 5e-3 and $\beta=(0.9, 0.999)$.
The results can be seen in \autoref{fig:image-colorization-demo}, where the model correctly identifies between neutral and non-neutral units, but has trouble determining whether a unit belongs to Player 1 or Player 2 due to both players having random starting corners of the map.

\end{document}